\documentclass[times, review, 10pt]{elsarticle}

\usepackage{amssymb}
\usepackage{amsmath}

\usepackage{multirow}
\usepackage{tabularx}
\usepackage{float}
\usepackage{makecell}
\usepackage{booktabs}
\usepackage{xcolor}
\usepackage{adjustbox}
\usepackage{subcaption}
\usepackage{graphicx}

\usepackage{algorithm}
\usepackage{algorithmic}
\usepackage{url}
\journal{Pattern Recognition}

\begin{document}

\begin{frontmatter}
\title{A Comprehensive Evaluation on Quantization Techniques for Large Language Models}





%
\author[1]{Yutong Liu}
\ead{2432031@tongji.edu.cn}
\author[1]{Cairong Zhao\corref{cor1}}
\ead{zhaocairong@tongji.edu.cn}
\author[2]{Guosheng Hu}
\ead{g.hu@bristol.ac.uk}

\address[1]{School of Computer Science and Technology, Tongji \\University, Shanghai 201804, China}
\address[2]{School of Engineering Math and Technology, University\\ of Bristol, BS8 1QU, UK}

\cortext[cor1]{Corresponding author.}
\begin{abstract}

For large language models (LLMs), post-training quantization (PTQ) significantly reduces memory and computational overhead. Despite rapid advancements, existing papers often lack experiments on the same ground. In addition, since quantization solutions typically contain multiple components, it is important to analyze the connections among numerous existing methods to gain an in-depth understanding of this field.

To bridge these gaps, we conduct an up-to-date and extensive review of the state-of-the-art methods and perform comprehensive evaluations of low-bit quantization on representative text-only LLMs on the same ground to ensure fair comparisons. To our knowledge, this fair and extensive investigation remains critically important yet significantly underexplored in the research community.

To better understand the connections, we first decouple published quantization into two steps: pre-quantization transformation to reduce outliers, and quantization error mitigation to offset performance loss. Second, we extensively evaluate the impact of quantization settings, including granularity and symmetry. Third, we analyze the performance of the latest MXFP4 and NVFP4 data formats.

Our experimental results first demonstrate that optimizing rotation and scaling yields the best pre-quantization transformation, while combining low-rank compensation with GPTQ occasionally outperforms using GPTQ alone. Second, we verify that finer granularity improves performance at the cost of storage and asymmetric quantization benefits activations significantly more than weights. Third, we find that the format and precision of scaling factors have a significant impact on performance; notably, rotation-based transformation is effective for INT4 but demonstrates little improvement for MXFP4 and NVFP4, inspiring further investigation in this field.

\end{abstract}





\begin{keyword}
Model Quantization \sep  Post-training Quantization \sep  LLM \sep Efficient AI 

\end{keyword}
\end{frontmatter}

\section{Introduction} 
Large language models (LLMs) have achieved remarkable performance in capturing complex semantic patterns across diverse tasks. Nonetheless, LLMs' inference process incurs significant memory and computational overhead, leading to high inference latency and posing challenges for computations on resource-limited GPUs. From the perspective of pattern representation, the high-dimensional weight distributions and activation outliers in LLMs make traditional compression techniques prone to significant information loss, thereby degrading the model's discriminative performance.

Quantization is a promising solution to reduce the memory and computational cost of LLMs, enhancing their inference efficiency. In recent years, quantization techniques have developed rapidly, including Quantization-Aware Training (QAT) methods \cite{Dettmers2023QLoRA,xu2024qalora,chen-etal-2025-efficientqat},  Post-Training Quantization (PTQ), and hybrid schemes that synergize quantization with pruning or distillation \cite{Qu_2025_CVPR,dettmers2024spqr}.
In this work, we focus on PTQ because it is more widely used than QAT in real applications. And our evaluation specifically targets text-only LLMs.

Additionally, LLMs' quantization has been studied in a wide spectrum of precision configurations. Representative settings include W4A4 \cite{zhao2024atom,shao2023omniquant,liu2024spinquant,hu2025ostquant}, W8A8 \cite{dettmers2022gpt3,xiao2023smoothquant,wei2023outlier} and W4A16 \cite{frantar2022gptq,chee2023quip,tseng2024quip2,lin2024awq} and mixed-precision settings such as W4A8 \cite{lin2024qserve}: W4A4 quantizes both weights and activations to 4-bit precision (INT4); W8A8 and W4A16 quantize weights to INT8 and INT4 and activations to INT8 and FP16/BF16 respectively;
Mixed-precision methods aim to improve accuracy by using higher bit-widths for parts that are more difficult to quantize \cite{zhao2024atom, li2023llm, huang2025slim, zhao2024poster,chenprogressive}. Additionally, non-uniform schemes utilize distribution-aware grids to minimize reconstruction loss in high-density regions \cite{dotzel2024learning, kim2024squeezellm, zhangleanquant}.

Among these three settings, W4A4 is currently the main focus of research, as it holds great potential for improving memory efficiency and inference speed. However, applying lossless W4A4 quantization to LLMs remains a challenging problem. To make our evaluation more cutting-edge and provide the research community with up-to-date insights, we select W4A4 as the primary precision configuration to conduct a comprehensive evaluation.

Many papers report breakthrough performance in quantization. We find that existing methods are often combinations of multiple techniques, since using a single technique alone typically fails to achieve excellent quantization performance. OSTQuant \cite{frantar2022gptq} and FlatQuant \cite{sun2024flatquant} employ multiple techniques to adjust the weights and activations, including rotation and scaling, and also apply GPTQ \cite{frantar2022gptq}, a gradient-based method for mitigating quantization error, to quantize the transformed weights. Besides, many techniques share similar transformations; for instance, Quip\# \cite{tseng2024quip2} and QuaRot \cite{ashkboos2024quarot} both use rotation; also, AWQ \cite{lin2024awq} and SmoothQuant \cite{xiao2023smoothquant} both use scaling. Analysis of the interconnections among diverse methods is critical for a thorough understanding of the field.

Besides, these quantization methods are often not directly comparable, as quantization often contains multiple configurations. It is important to evaluate these methods on the same ground. However, there is a distinct lack of comprehensive evaluations on recent W4A4 PTQ methods. Early evaluations on quantization \cite{li2024evaluating,huang2024good} tend to assess methods individually under their respective original settings without exploring how different components interact when combined. While existing surveys provide overviews, they often lack both proper categorization and a comprehensive, fair evaluation. Merely analyzing and categorizing previous literature is insufficient for a deeper understanding.

To bridge these gaps, in this paper, we conduct an up-to-date and extensive review of state-of-the-art methods alongside a comprehensive, unified evaluation. On one hand, our work incorporates a structured classification and analysis to help readers better understand the literature and ground our evaluation. On the other hand, moving beyond simple categorization, our extensive evaluation benchmarks and synthesizes these approaches under a unified setting, offering a level of insight that prior surveys have overlooked.

We mainly investigate three dimensions. First, to achieve a better understanding of existing methods and make our evaluation on the same ground, we decompose the quantization process into two key steps: pre-quantization transformation and quantization error mitigation. The pre-quantization transformation step refers to preprocessing the data before quantization in order to mitigate the effect of outliers, making the data distribution smoother and more quantization-friendly. The quantization error mitigation step involves applying techniques to compensate for the errors introduced by quantization, thereby reducing the overall quantization error and improving performance. For each step, we select several state-of-the-art techniques and conduct an evaluation of the effectiveness of their combinations. 

Second, quantization involves a variety of configurations, such as granularity and whether symmetric or not. Different settings can lead to different performance outcomes and often involve trade-offs between efficiency and accuracy. For different precision levels, the optimal quantization settings may vary. And existing solutions usually do not use exactly the same setting, leading to unfair comparisons. To bridge this gap, we conduct a comprehensive evaluation on the same quantization configurations and provide recommendations on selecting appropriate configurations. And we integrate the methods used in this paper into our code library to facilitate performance evaluation. 

Third, NVIDIA’s GeForce RTX 50 Series, which is newly released and based on the Blackwell architecture, includes support for the FP4 data format (\textit{e.g.}, NVFP4 and MXFP4) in its Tensor Cores, showing significant advantages over INT4. However, the research on FP4 has not been well explored. To bridge this gap, we conduct research on weight-activation FP4 quantization under our two-step quantization framework in INT4, showing the great timeliness of our research. For FP4 quantization, we also observe that the format and precision of scaling factors have a significant impact on performance.

\section{Review of Existing Quantization Methods}

In this section, we first review the evolution of LLM quantization in Section \ref{sec:evolution}. In Section \ref{sec:two-steps}, motivated by the trend toward multi-component combinations, we organize recent quantization algorithms around a two-step decomposition: pre-quantization transformation and quantization error mitigation. Table \ref{tab:intromethod} shows our decomposition of the popular quantization methods. Section \ref{sec:settings} focuses on settings, including symmetry and granularity of quantization methods. Table \ref{tab:introconfig} summarizes the settings of existing methods, showing that they are not compared on the same ground. Section \ref{sec:fp4} discusses FP4 quantization and its variants, including MXFP4 and NVFP4.

\subsection{Evolution of LLM Quantization}
\label{sec:evolution}
LLM quantization has evolved along three parallel directions targeting different components of the inference pipeline, with weight-activation quantization undergoing the most dramatic shift from single techniques to multi-component combinations, the central insight motivating our two-step decomposition.

\textbf{Weight-only quantization: compressing memory footprint.}
Early LLM quantization focused on weight compression to mitigate memory bandwidth bottlenecks, establishing per-group quantization (GPTQ~\cite{frantar2022gptq}, AWQ~\cite{lin2024awq}) as the paradigm for W4A16 and W3A16 settings. Subsequent works pursued more aggressive compression: AQLM~\cite{egiazarian2024extreme} introduced multi-codebook additive quantization for 2-bit weights; SqueezeLLM~\cite{kim2024squeezellm} and LeanQuant~\cite{zhangleanquant} employed non-uniform quantization, with SqueezeLLM integrating dense-and-sparse representations; and QuIP\#~\cite{tseng2024quip2} leveraged Hadamard incoherence processing with lattice codebooks. Alternatively, residual errors were reconstructed via SVD decomposition~\cite{yao2023zeroquant,saha2024compressing,zhang2024qera}. Recently, ultra-low-bit methods pushed weight precision to the extreme: PT-BitNet~\cite{guo2025pt} scales 1-bit LLMs through post-training quantization; STBLLM~\cite{dong2025stbllm} breaks the 1-bit barrier via N: M structured sparsity with significance-aware grouping; ARB-LLM~\cite{li2025arb} iteratively refines binarization parameters; HBLLM~\cite{chen2026hbllm} incorporates Haar wavelet decomposition; and LittleBit~\cite{lee2026littlebit} achieves sub-1-bit boundaries via latent matrix factorization.


\textbf{Weight-activation quantization: from single techniques to multi-component combinations.}
Weight-activation quantization reduces both memory and computational cost by quantizing weights and activations together. At W8A8, activation outliers are manageable through simple transformations: SmoothQuant~\cite{xiao2023smoothquant} applies scaling alone to ease activation quantization, while Outlier Suppression+~\cite{wei2023outlier} combines shifting with scaling. At W4A4, the ability for 4-bit representation is narrow, which means the outlier problem is harder to handle and some W8A8 techniques are insufficient. For pre-quantization transformation, OmniQuant~\cite{shao2023omniquant} optimizes shifting and scaling, while subsequent methods leverage rotations to handle outliers: QuaRot~\cite{ashkboos2024quarot} employs random rotation, SpinQuant~\cite{liu2024spinquant} learns it, and DuQuant~\cite{lin2024duquant} introduces dual rotation with zigzag permutation. The field then evolved toward jointly optimizing multiple transformations: QServe~\cite{lin2024qserve} combined scaling and rotation. FlatQuant~\cite{sun2024flatquant} and OSTQuant~\cite{hu2025ostquant} jointly optimized rotation and scaling matrices. For quantization error mitigation, weight-only techniques were also adopted into weight-activation combinations: GPTQ~\cite{frantar2022gptq}, originally designed for weight-only quantization, was widely used as a component for weight quantization in W4A4 methods. To mitigate errors, some works~\cite{zhang2024lqer,fu2025qwt} incorporate additive branches; CBQ~\cite{ding2025cbq} addresses intra- and inter-layer discrepancies via cross-block reconstruction with LoRA-Rounding; BoA~\cite{kim2025boa} and TurboBoA~\cite{TurboBoA} introduce backpropagation-free attention-aware quantization; and Qronos~\cite{zhang2026qronos} explicitly corrects accumulated errors from previously quantized layers. At the representation level, hardware support from NVIDIA Blackwell Tensor Cores for FP4 formats introduces new considerations: MXFP4~\cite{rouhani2023microscaling,ocp2024mx} and NVFP4~\cite{nvfp4} leverage block-level shared exponents; AMXFP4~\cite{lee2024amxfp4} introduces asymmetric microscaling to suppress outliers; and Any4~\cite{elhoushi2025any4} learns data-driven numerical representations rather than using fixed formats. Beyond standard Transformers, MambaQuant~\cite{xumambaquant} presents a PTQ framework for Mamba/SSM via variance-aligned rotation, while MxMoE~\cite{duanmu2025mxmoe} quantizes MoEs using sensitivity- and frequency-driven mixed precision.

\textbf{KV cache quantization: compressing context memory.} As weight-activation quantization reduces the cost of linear computation, the KV cache emerges as the dominant memory bottleneck during long-context inference, motivating a parallel line of research. Recent methods address KV cache outliers through per-channel key quantization with sensitivity-weighted non-uniform codes~\cite{hooper2024kvquant}, asymmetric key-value quantization with different granularity for keys and values~\cite{liu2024kivi}, sensitivity-aware layer-wise mixed-precision configuration~\cite{likvtuner}, polar coordinate decoupling of magnitude and phase~\cite{wu2026polarquant}, and attention-guided preservation of important KV pairs~\cite{liu2024intactkv}. Recent works further explore progressive mixed precision for long-CoT~\cite{liu2026pm} and channel-aware mixed precision for long-context inference~\cite{liao2026channel}.

From this evolution, we find that the quantization methods can be decomposed into two steps: pre-quantization transformation and quantization error mitigation, as detailed in Section~\ref{sec:two-steps}, which provides a better perspective for understanding LLM quantization.
\begin{table}[h]
\centering
\begin{tabular}{c lcl lcl} 
\toprule 
Methods && \makecell{Pre-quantization \\ Transformation} &&& \makecell{Quantization Error \\ Mitigation} &\\

\midrule 
SmoothQuant \cite{xiao2023smoothquant} && Scaling &&& RTN &\\
Outlier Suppression+ \cite{wei2023outlier}                 && Shifting + Scaling &&& RTN  & \\
GPTQ      \cite{frantar2022gptq}        && - &&& GPTQ &\\
ZeroQuant-v2 \cite{yao2023zeroquant}  && - &&& Low-rank &\\
QuIP  \cite{chee2023quip}             && Rotation &&& GPTQ &\\ 
AWQ       \cite{lin2024awq}            && Scaling &&& RTN  &\\  
Atom  \cite{zhao2024atom}               && Reorder &&& GPTQ &\\
OmniQuant \cite{shao2023omniquant}     && Shifting+Scaling&&& GPTQ &\\
Qserve \cite{lin2024qserve}           && Scaling + Rotation &&& GPTQ &\\
LQER      \cite{zhang2024lqer}          && Scaling &&& Low-rank &\\
QuaRot  \cite{ashkboos2024quarot}      && Rotation &&& GPTQ &\\
SpinQuant \cite{liu2024spinquant}      && Rotation &&& GPTQ &\\
ResQ    \cite{saxena2024resq}          && Reorder + Rotation &&& GPTQ &\\
FlatQuant \cite{sun2024flatquant}       && Scaling + Rotation&&& GPTQ &\\
OSTQuant  \cite{hu2025ostquant}      && Scaling + Rotation &&& GPTQ &\\
\bottomrule 
\end{tabular}
\caption{Two-step decomposition of quantization methods.}\label{tab:intromethod}
\end{table}
\begin{table}[t]
\centering
\begin{tabular}{c c lccl c} 
\toprule 
\multirow{2}*{Methods} &  \multirow{2}*{Precision} & & \multicolumn{2}{c}{Symmetric} & & \multirow{2}*{Granularity}  \\ 

\cline{4-5} 
& && W & A &&   \\

\midrule 
SmoothQuant \cite{xiao2023smoothquant} &  W8A8 && T & F && T,C  \\
GPTQ \cite{frantar2022gptq}       &  W3/4A16 && T & - && G  \\
ZeroQuant-v2 \cite{yao2023zeroquant}   &   W2/3/4A16,W4A8 && T,F & T,F && G   \\ 
QuIP \cite{chee2023quip}             &  W2/3/4A16 && T & - && G  \\ 
AWQ \cite{lin2024awq}            &   W3/4A16 && T & - && G  \\ 
Atom \cite{zhao2024atom}              &  W3/4A3/4 && T & F && G  \\
OmniQuant \cite{shao2023omniquant}     &  W2/3/4A16,W4/6/8A4/6/8  && F & F && C \\
Qserve \cite{lin2024qserve}            &  W4A8 && T & F && C\\ 
LQER \cite{zhang2024lqer}         &  W4A6/8 && T & F && G\\ 
QuaRot  \cite{ashkboos2024quarot}      &  W4A4 && T & F && C,G\\
SpinQuant \cite{liu2024spinquant}      &  W4A4/8 && T & F && C\\
ResQ    \cite{saxena2024resq}          &  W4A4 && T & F && C \\
FlatQuant \cite{sun2024flatquant}      &  W4A4 && T & F && C \\
OSTQuant  \cite{hu2025ostquant}        &  W4A4 && T & F && C  \\ 
\bottomrule 
\end{tabular}
\caption{Symmetry and granularity of quantization methods. In granularity, T stands for per-tensor, C represents per-channel or per-token, and G means per-group.}
\label{tab:introconfig}
\end{table}

\subsection{Two steps of quantization}
\label{sec:two-steps}
\subsubsection{Pre-quantization transformation}
\label{sec:dn}
\begin{figure}[ht]
\centering
\includegraphics[width=0.9\textwidth]{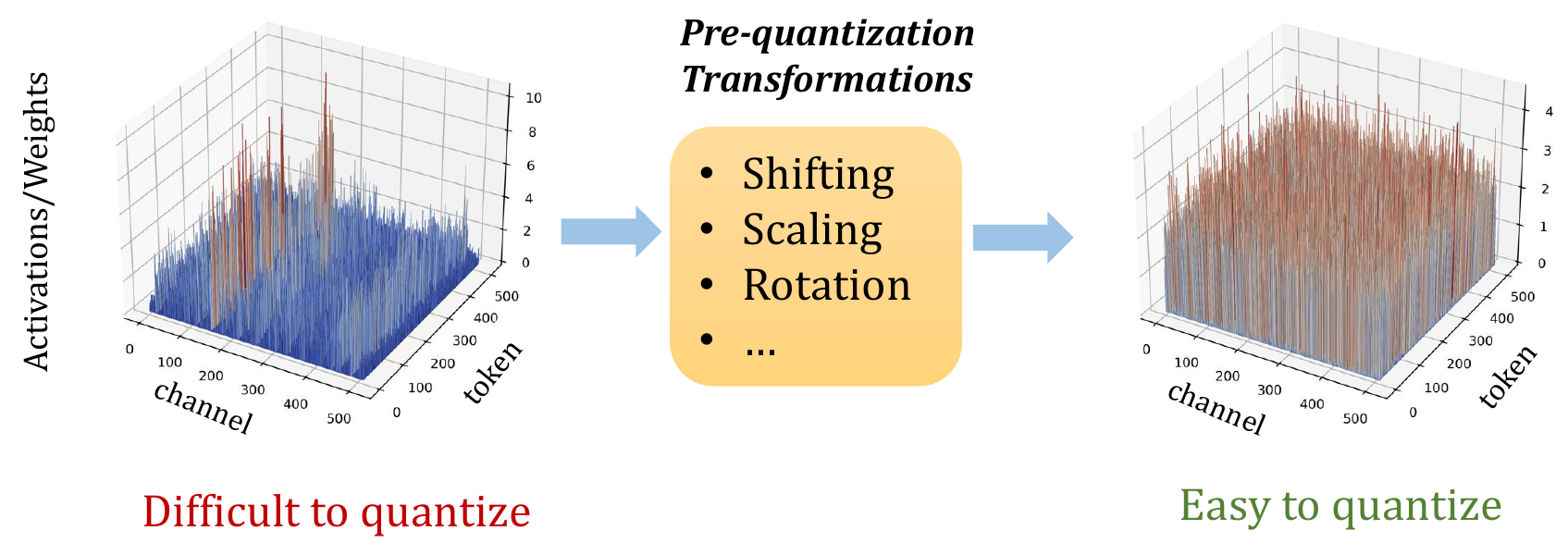}
\caption{Pre-quantization transformation: a process that transforms activations and weights to make them easier to quantize. The left graph illustrates a tensor that is difficult to quantize due to the presence of numerous outliers, while the right graph shows that after the pre-quantization transformation, the tensor becomes more uniformly distributed, making quantization easier.}
\label{fig:data}
\end{figure}

Pre-quantization transformation is a step before quantization, applying operations such as shifting, scaling, and rotation to process weights and activations. These transformations aim to flatten and smooth the hard-to-quantize tensors, thereby enhancing the quantization performance. LLMs are difficult to quantize due to the outliers in the activations. Compared to most of the activation values, the magnitude of outliers is substantially higher, often exceeding 100x larger in some layers. The outliers will dominate the maximum magnitude measurement, leading to low effective quantization levels. This pre-quantization transformation step is illustrated in Fig. \ref{fig:data}.
\paragraph{Shifting}Activation distribution is often asymmetric, and this asymmetry varies across different channels. Aligning the center of each channel can improve the performance of quantization, since we quantize activations per-token rather than along the channel dimension. Outlier Suppression+ \cite{wei2023outlier} proposes shifting activation values to eliminate the asymmetry feature across channels. The shifting process uses a vector $T$ that transfers the channels of the activation. For a linear layer, it is formulated as:
\begin{equation}
Y=XW+B= (\underbrace{(X-T)}_{\hat{X}})W+(\underbrace{TW+B}_{\hat{B}})
\end{equation}
where the layer input $X\in \mathbb{R}^{T\times C_{in}}$ is shifting and becomes $\hat{X}$. $T\in \mathbb{R}^{1\times C_{in}}$ is a vector representing the channel-wise shifting parameters. To keep mathematical equivalence, an additional term, the product of $T$ and $W\in \mathbb{R}^{C_{in}\times C_{out}}$, is added to the bias $B\in \mathbb{R}^{1\times C_{out}}$. 

\begin{table}[ht]
\centering
\begin{tabular}{ccc} 
\toprule 
\textbf{Methods} & \textbf{Shifting Strategy} & \textbf{Shifting parameters $T$} \\ 
\midrule
OS+ \cite{wei2023outlier} & Pre-scaling shifting
& 
 $T_i = \frac{\min(X_i) + \max(X_i)}{2} $\\
OmniQuant \cite{shao2023omniquant} & Pre-scaling shifting
& Gradient-based optimization \\

\bottomrule
\end{tabular}
\caption{Comparison of shifting-based quantization methods.}
\label{tab:shifting_methods}
\end{table}

As shown in Table \ref{tab:shifting_methods}, Outlier Suppression+ \cite{wei2023outlier} uses calibration to get the vector $T$, while OmniQuant \cite{shao2023omniquant} employs learnable channel-wise shifting for outlier suppression, combined with the learnable scaling technique discussed in the next paragraph, and jointly optimizes these two kinds of pre-quantization transformation parameters.

\paragraph{Scaling}
\label{sec:scale}

Scaling is a commonly used pre-quantization transformation method, which scales activations in different input channels before quantization, and inversely scales the corresponding weight channels to make the results stay unchanged.
SmoothQuant \cite{xiao2023smoothquant} first proposes using scaling to smooth the quantization difficulty from activations to weights. Compared to weights, activations have more outliers that make quantization harder. It is observed that outliers in activations appear in certain channels, so applying channel-wise scaling to activations can eliminate the outliers to a certain degree, using the following equation for linear layers: 
\begin{equation}
Y=XW=( \underbrace{X \cdot S^{-1}}_{\hat{X}}) ( \underbrace{S \cdot W}_{\hat{W}} ) 
= \hat{X} \hat{W}
\end{equation}
where $S$ is a diagonal matrix composed of scaling factors of different channels, which can be obtained in various ways. 
\begin{table}[ht]
\centering
\renewcommand{\arraystretch}{0.8}
\begin{tabular}{ccc} 
\toprule 
\textbf{Methods} & \textbf{Scaling Strategy
} & \textbf{Scaling parameters $S$} \\ 
\midrule
SmoothQuant \cite{xiao2023smoothquant} &Scaling only
& 
$ \makecell{S_i = \frac{\max(|X_i|)^{\alpha}}{\max(|W_i|)^{1-\alpha}}}$\\

 AWQ \cite{lin2024awq} & Scaling only

&  $ \makecell{S_i=|X_i|_{avg}^\alpha \\  \alpha^*= \arg\min_\alpha \mathcal{L}(S_{Xi}^\alpha)}$ \\

OS+ \cite{wei2023outlier} & Post-shifting scaling

&$ \makecell{S_i = \max\left(1.0, \frac{\max(X_i - T_i)}{t}\right) \\  t^*=\arg\min_t \mathcal{L}(S_i) } $ \\
OmniQuant \cite{shao2023omniquant} & Post-shifting scaling

& Gradient-based optimization \\
 FlatQuant \cite{sun2024flatquant} & Pre-rotation scaling

& \makecell{Calibration initialization \\and gradient-based optimization }\\
 OSTQuant \cite{hu2025ostquant} &Post-rotation scaling

& Gradient-based optimization \\
\bottomrule
\end{tabular}
\caption{Comparison of scaling-based quantization methods.}
\label{tab:scaling_methods}
\end{table}

As shown in Table \ref{tab:scaling_methods}, SmoothQuant \cite{xiao2023smoothquant} uses calibration to compute $S$. AWQ \cite{lin2024awq}, as a weight-only quantization, approaches the problem from the perspective of protecting salient weights. Its scaling parameter is determined by the average magnitude of the activation, while the optimal exponent $\alpha^*$ is found by minimizing the loss through grid searching. Outlier Suppression+ \cite{wei2023outlier} also uses grid searching optimization to find the optimal scaling after applying shifting.  OmniQuant \cite{shao2023omniquant} optimizes the quantization parameters via gradient-based methods to minimize the block-wise quantization error. FlatQuant \cite{sun2024flatquant} and OSTQuant \cite{hu2025ostquant} both use learnable per-channel scaling, but at different stages of pre-quantization transformation, leading to different practical meanings. 

\paragraph{Rotation}
\label{sec:rotation}
The rotation operation is a technique that adjusts data distribution by multiplying the data matrix by an orthogonal matrix. This operation largely changes the activation distributions, reduces outliers in the data, and thereby enhances quantization accuracy. The rotation process in a linear layer can be formulated as:
\begin{equation}
Y=XW=(\underbrace{XO}_{\hat{X}}) (\underbrace{O^\top W}_{\hat{W}}) 
= \hat{X} \hat{W}
\end{equation}
where the input $X$ is rotated with an orthogonal matrix $O$, resulting in $\hat{X}=XO$. To preserve the model's output, the weight $W$ is simultaneously transformed by $O^\top$, which is the transpose of $O$. 

\begin{table}[ht]
\centering
\renewcommand{\arraystretch}{0.8}
\begin{tabular}{ccc} 
\toprule 
\textbf{Methods} & \textbf{Rotation Strategy} & \textbf{Rotation parameters $O$} \\ 
\midrule
QuIP \cite{chee2023quip} & Rotation only
& Random orthogonal matrices \\

QuIP\# \cite{tseng2024quip2}& Rotation only
& Random Hadamard matrices \\

QuaRot \cite{ashkboos2024quarot} & Rotation only
& Random Hadamard matrices \\

SpinQuant \cite{liu2024spinquant} & Rotation only
& Gradient-based optimization \\

FlatQuant \cite{sun2024flatquant} & Post-scaling Rotation
& Gradient-based optimization \\

OSTQuant \cite{hu2025ostquant} & Pre-scaling Rotation
& Gradient-based optimization\\
\bottomrule
\end{tabular}
\caption{Comparison of rotation-based quantization methods.}
\label{tab:rotation_methods}
\end{table}
As shown in Table \ref{tab:rotation_methods}, QuIP \cite{chee2023quip} first introduced rotation with random orthogonal matrices to reduce incoherence in weight-only quantization. QuIP\# \cite{tseng2024quip2} replaces this with random Hadamard matrices for greater efficiency. QuaRot \cite{ashkboos2024quarot} extends rotation to both weights and activations quantization using random Hadamard matrices, effectively eliminating outliers. SpinQuant \cite{liu2024spinquant} further improves stability by optimizing the rotation matrix on the Stiefel manifold to minimize quantization-induced performance variance, while still defaulting to random Hadamard for a few online rotations.

Rotation is a highly effective pre-quantization transformation technique for reducing outliers in low-bit quantization, and it has also been widely adopted in many other quantization methods together with other pre-quantization transformation techniques to further improve performance  (\textit{e.g.}, ResQ \cite{saxena2024resq}, QServe \cite{lin2024qserve}, FlatQuant \cite{sun2024flatquant}, and OSTQuant \cite{hu2025ostquant}, etc).

\subsubsection{Quantization error mitigation}
\label{sec:er}
\begin{figure}
\centering
\includegraphics[width=\textwidth]{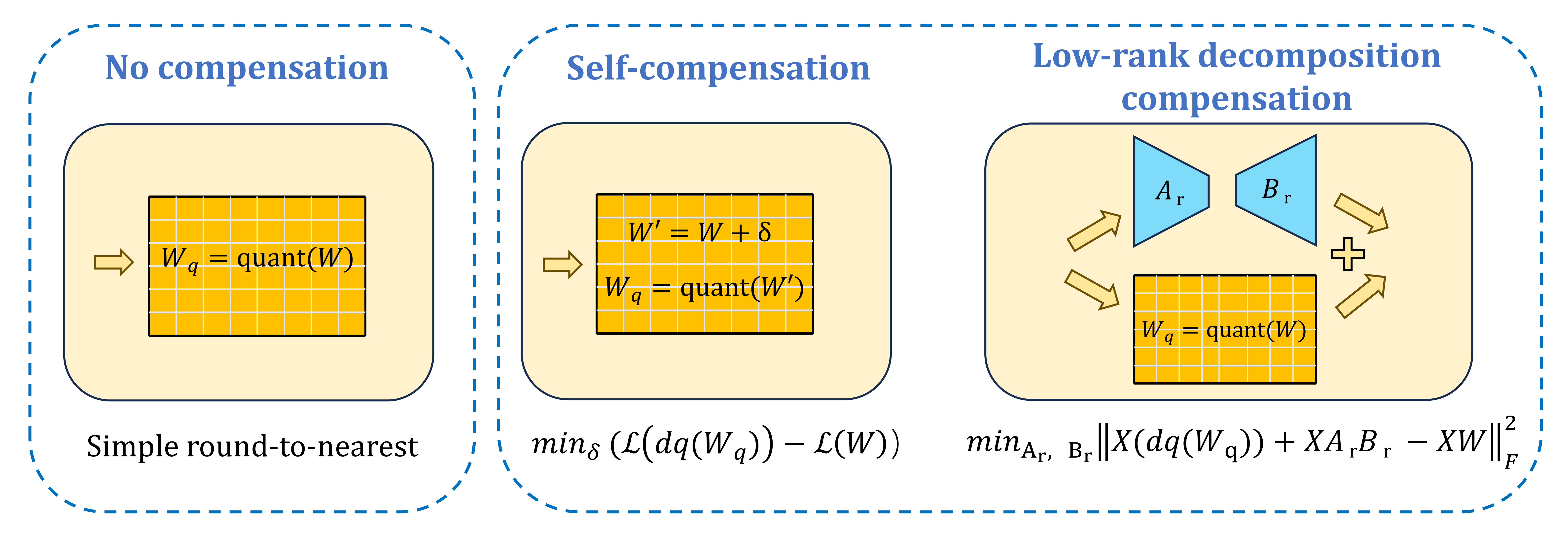}
\caption{Quantization error mitigation: compensate the error produced by quantization. Here, $dq(W_q)$ denotes dequantizing  $W_q$ back to the original precision value space, for computational purposes. }
\label{fig: error}
\end{figure}

The quantization error mitigation step employs techniques that mitigate the error introduced by quantization by compensating for it. This error between the original model and the quantized model can be computed from calibration datasets, and compensated to the original weights in linear layers or through low-rank branches. Fig. \ref{fig: error} demonstrates two commonly used quantization error mitigation methods. 
\paragraph{Self-compensation} 
We define self-compensation as directly compensating for errors on the original weights. Existing methods that employ self-compensation are generally based on analyzing the loss error introduced by quantization and make use of its Hessian-based  quadratic approximation, which can be formulated as:
\begin{equation}
\begin{aligned}
\Delta \mathcal{L}
&= \mathcal{L}(dq(W_q)) - \mathcal{L}(W) \\
&\approx \frac{1}{2}
\bigl(dq(W_q) - W\bigr)^\top
H
\bigl(dq(W_q) - W\bigr)
\end{aligned}
\label{eq:epsilon}
\end{equation}
where $\mathcal{L}(W)$ is the loss function of the model evaluated at weights $W$. $W_q$ and $dq(W_q)$ denote the quantized weights and their dequantized version, respectively. Self-compensation methods need to find an optimal weight perturbation $\delta $ to add to $W$, such that the quantization loss error in Eq. (\ref{eq:epsilon}) is minimized. Table \ref{tab:self-compensation} summarizes the computation strategies of $\delta $ adopted by various self-compensation quantization methods.

\begin{table}[ht]
\centering   
\renewcommand{\arraystretch}{0.8}
\begin{tabular}{cc}
\toprule 
\textbf{Methods} & \textbf{Weight perturbation $\delta$ computation} \\ 
\midrule
OBQ \cite{frantar2022optimal} &Layer-wise MSE \\
GPTQ \cite{frantar2022gptq} &Layer-wise MSE \\
QuIP \cite{chee2023quip}&LDL decomposition \\
GPTAQ \cite{li2025gptqv2} &Consider input deviation\\ 
GQuant \cite{kim2025guidedquant} &Add gradient information \\
\bottomrule
\end{tabular}
\caption{Comparison of self-compensation quantization methods.}
\label{tab:self-compensation}
\end{table}

Some methods simplify the quantization problem to minimize the squared error between the quantized and original output of each layer \cite{frantar2022optimal,frantar2022gptq}.
Based on the squared error loss function, OBQ \cite{frantar2022optimal} and GPTQ \cite{frantar2022gptq} compute the Hessian matrix used in Eq. (\ref{eq:epsilon}) and further calculate the optimal weight perturbation $\delta$. Unlike OBQ, which uses greedy orders to quantize weights, GPTQ quantizes all rows of weights in a fixed order to achieve a faster quantization process. 

Some other methods take into account additional factors beyond the linear layer-wise formulation. While also aiming to minimize the quadratic proxy of the weight error in Eq. (\ref{eq:epsilon}), QuIP \cite{chee2023quip} performs an LDL decomposition of the Hessian matrix to construct a compensation term that corrects for quantization-induced errors. GPTAQ \cite{li2025gptqv2}, built upon GPTQ, addresses the limitation that GPTQ’s layer-wise optimization objective ignores the deviation between a layer’s actual input and that under the full-precision model. GQuant \cite{kim2025guidedquant} further observes that the layer-wise MSE treats all hidden features equally, and therefore incorporates per-feature gradients into the objective function to improve quantization accuracy.

\paragraph{Low-rank compensation}

Beyond compensating errors directly on the original weights, an alternative approach is to introduce an auxiliary branch dedicated to quantization error mitigation. 

Some prior works leverage a low-rank term to reconstruct quantization error (\textit{e.g.}, ZeroQuant-v2 \cite{yao2023zeroquant}, CALDERA \cite{saha2024compressing}, LQER \cite{zhang2024lqer}, QERA \cite{zhang2024qera}, etc), where the low-rank branch can either be quantized to higher precision, such as INT8, or remain in FP16.
After incorporating a low-rank branch, the output $Y$ is computed as:
\begin{equation}
Y=Xdq(W_q)+XAB
\label{eq:y=xw+xab}
\end{equation}
where $A \in \mathbb{R}^{C_{in} \times r}$, $B \in \mathbb{R}^{r \times C_{out}}$ are low-rank matrices, and $r \ll \min(d, k)$ is the bottleneck rank. The product $AB \in \mathbb{R}^{d \times k}$ serves as a low-rank approximation of the quantization error. Table \ref{tab:lowrank-compensation} illustrates the computation process of the low-rank terms in some quantization methods.

\begin{table}[ht]
\centering
\renewcommand{\arraystretch}{0.8}
\begin{tabular}{cc}
\toprule 
\textbf{Methods} & \textbf{Low rank terms $A, B$ computation} \\
\midrule
   ZeroQuant-v2 \cite{yao2023zeroquant}& $\text{SVD} (dq(W_q)-W)$\\ [4pt]
LQER \cite{zhang2024lqer}&  $\text{SVD} (dq(W_q)-W ) $\\ [4pt]
L$^2$QER \cite{zhang2024lqer}&  $\text{SVD}(S(dq(W_q)-W)), \space    A :=S^{-1}A  $\\ [4pt]
CALDERA \cite{saha2024compressing}&  LPLR factorization \cite{saha2024compressing} \\  [4pt]
QERA \cite{zhang2024qera}&  $\text{SVD} (R^\frac{1}{2}_{X\!X}(dq(W_q)-W)),\space A :=(R^\frac{1}{2}_{X\!X})^{-1}A $ \\ [4pt]
\bottomrule
\end{tabular}
\caption{Comparison of low-rank compensation quantization methods. $A, B$ represent the low-rank terms. $R_{X\!X}$ denotes the autocorrelation matrix of the input $X$.}
\label{tab:lowrank-compensation}
\end{table}

ZeroQuant-v2 \cite{yao2023zeroquant} and LQER \cite{zhang2024lqer} apply SVD to reconstruct the quantization error matrix $E=W-dq(W_q)$. Based on LQER, L$^2$QER, which is proposed in LQER, scales the $E$ before applying SVD heuristically, and inversely scales $A_k$. This operation aims to more precisely approximate the quantization error corresponding to the salient weights.

In addition to the aforementioned approaches, CALDERA \cite{saha2024compressing} and QERA \cite{zhang2024qera} derive the optimal low-rank terms for quantization error mitigation by framing it as an optimization problem that minimizes the squared error between each layer's output in Eq. (\ref{eq:y=xw+xab}) and its original output.

In contrast to approaches that introduce low-rank branches externally, QwT \cite{fu2025qwt} integrates lightweight linear layers within each decoder layer block of LLMs, directly compensating quantization errors, offering new insights for quantization error mitigation.

\subsection{Settings of quantization}
\label{sec:settings}
\subsubsection{Symmetric vs. asymmetric}
\label{sec:sym}
\begin{figure}[H]
  \centering
  \begin{subfigure}[b]{0.7\textwidth}
    \includegraphics[width=\textwidth]{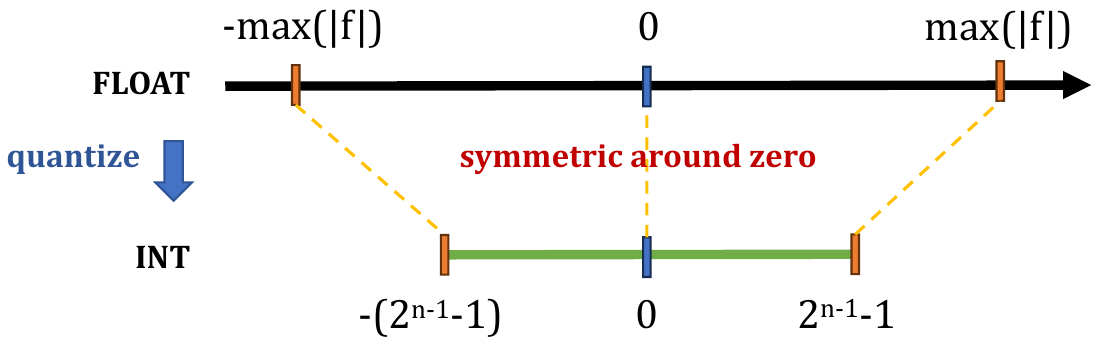}
    \caption{Symmetric quantization}
  \end{subfigure}
  \hfill 
  \begin{subfigure}[b]{0.7\textwidth}
    \includegraphics[width=\textwidth]{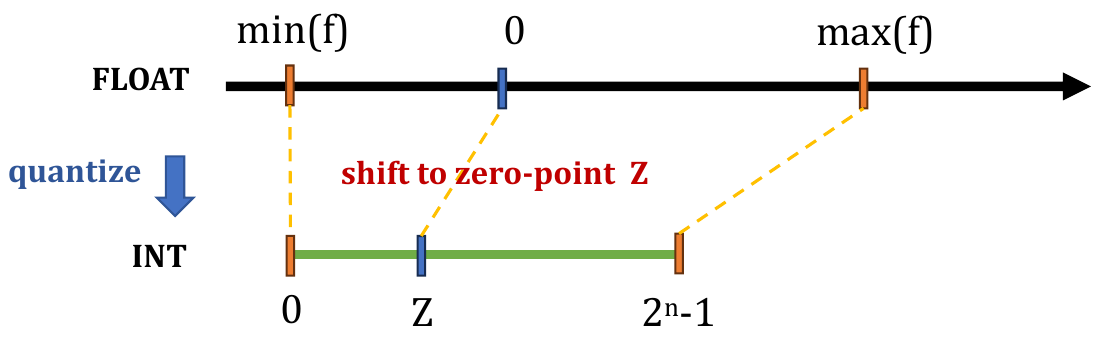}
    \caption{Asymmetric quantization}
  \end{subfigure}
  \caption{Symmetry of quantization}
  \label{fig:symmetric}
\end{figure}

Uniform quantizers can be broadly categorized into symmetric and asymmetric schemes, depending on how the zero point is handled during the mapping from floating-point to integer values, as illustrated in Fig. \ref{fig:symmetric}. 
Symmetric quantization forces the zero point to align with the floating point, resulting in a quantization range that is symmetric with respect to zero. In contrast, asymmetric quantization dynamically shifts the zero point according to the data range, enabling a tighter fit to skewed distributions at the cost of additional computational complexity.

Symmetric quantization enables a very simple dequantization process. For linear layers, dequantization can be efficiently performed by multiplying the quantized output by the product of the scale factors for the weights and activations.
This simplifies computation and is often preferred in hardware implementations. However, symmetric quantization may lead to suboptimal representation when the data distribution is not centered around zero.

Asymmetric quantization incurs additional computational cost during dequantization, while it often outperforms symmetric quantization in terms of accuracy. 

As such, the selection of quantization symmetry should be made with consideration of the specific use case and performance-complexity trade-offs.

\subsubsection{Granularity of quantization}
\label{sec:gran}
\begin{figure}
\centering
\includegraphics[width=0.75\linewidth]{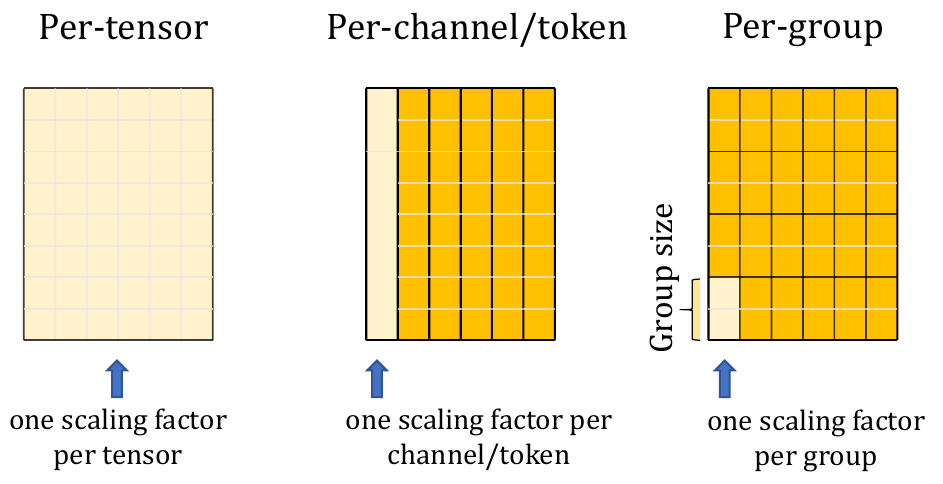}
\caption{Granularity of quantization: (1) Per-tensor quantization; (2) Per-channel/token quantization; (3) Per-group quantization. }
\label{fig:threegranu}
\end{figure}

The granularity of quantization often largely affects the low-bit quantization performance. Commonly adopted granularities include per-tensor, per-channel, or per-token, and per-group, as illustrated in Fig. \ref{fig:threegranu}. 
In per-tensor quantization, a single set of affine transformation parameters is applied to the entire tensor, with the scale ($\Delta$) and zero point ($Z$) computed from the tensor's minimum and maximum values. A finer-grained alternative is per-channel quantization for weights and per-token quantization for activations, where each input channel of the weights or each token of the activations shares the same quantization parameters. 
For low-bit quantization scenarios, even finer granularity, per-group quantization, is often employed to further enhance performance. In this case, the elements within a single channel or token are subdivided into groups of a predefined size, with each group quantized independently using its own parameters.

Finer granularity typically leads to improved performance but also incurs higher memory overhead due to the need to store more quantization parameters. Besides, as QServe \cite{lin2024qserve} mentioned, per-group quantization will also raise computational cost at the hardware level. Therefore, while finer granularity can enhance accuracy, it may also reduce the overall efficiency of the system.

In weight-only quantization, papers such as GPTQ \cite{frantar2022gptq}, AWQ \cite{lin2024awq}, and QuIP \cite{chee2023quip} all employ per-group quantization to further enhance performance. For weight-activation quantization, Atom \cite{zhao2024atom} also adopts a per-group quantization strategy. However, with the introduction of rotation-based methods in QuaRot \cite{ashkboos2024quarot} and SpinQuant \cite{liu2024spinquant}, the performance of per-channel quantization has been significantly improved. Inspired by this, subsequent papers like FlatQuant \cite{sun2024flatquant}, ResQ \cite{saxena2024resq}, and OSTQuant \cite{hu2025ostquant} have also adopted per-channel granularity for quantization. 
\subsection{FP4 Quantization and Variant Formats}
\label{sec:fp4}
Recently, NVIDIA's GeForce RTX 50 Series based on NVIDIA's Blackwell architecture \cite{nvidia2024blackwell} has been released, which has enabled very low-precision operations using FP4 variants like MXFP4 \cite{rouhani2023microscaling,ocp2024mx} and NVFP4 \cite{nvfp4}. 
Both MXFP4 and NVFP4 store each value in an E2M1 format, which consists of one sign bit, one mantissa bit, and two exponent bits. Unlike the INT4 format, they adopt a fixed granularity and use symmetric per-group quantization, with MXFP4 employing a group size of 32 and NVFP4 using a finer granularity of 16. Each group is associated with a scaling factor, and this rescaling process is hardware-accelerated. MXFP4 uses an FP8 (E8M0) format, while NVFP4 uses an FP8 (E4M3) format. Additionally, NVFP4 incorporates an extra FP32 scaling factor per tensor to extend its representable range.

The value of an encoding FP4 (E2M1) data is represented as:
\begin{equation}
\text{value} =
\begin{cases}
(-1)^S \times 2^{1-\text{bias}} \times (0 + 2^{-m} \times M) & \text{if } E = 0  \\
(-1)^S \times 2^{E-\text{bias}} \times (1 + 2^{-m} \times M) & \text{if } E > 0
\end{cases}
\end{equation}
where $S, E, M$ correspond to the sign, exponent, and mantissa fields, respectively. The bias is the exponent bias, which is set to 1. So the FP4 (E2M1) encoding values are [-6, -4, -3, -2, -1.5, -1, -0.5, 0, 0.5, 1, 1.5, 2, 3, 4, 6].

\begin{figure}
  \centering
  \captionsetup[subfloat]{font=footnotesize, labelfont=footnotesize}
  \subfloat[]{\includegraphics[width=0.34\textwidth]{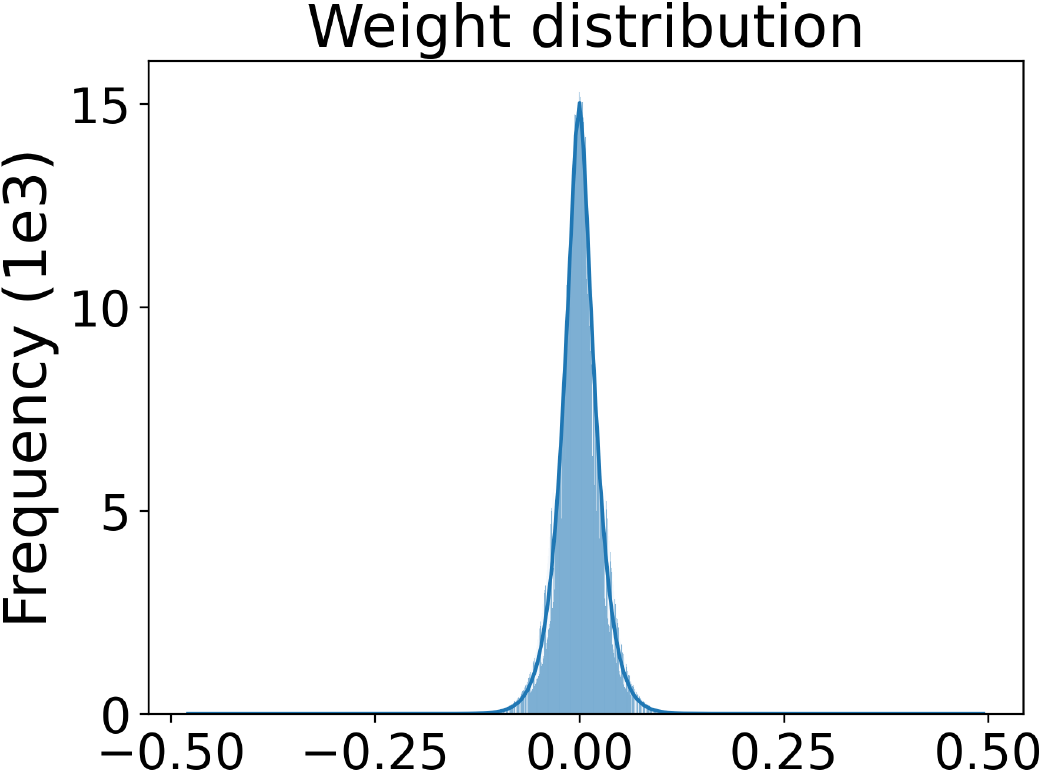}%
    \label{fig:sub1}}
    \hfil %
      \subfloat[]{\includegraphics[width=0.34\textwidth]{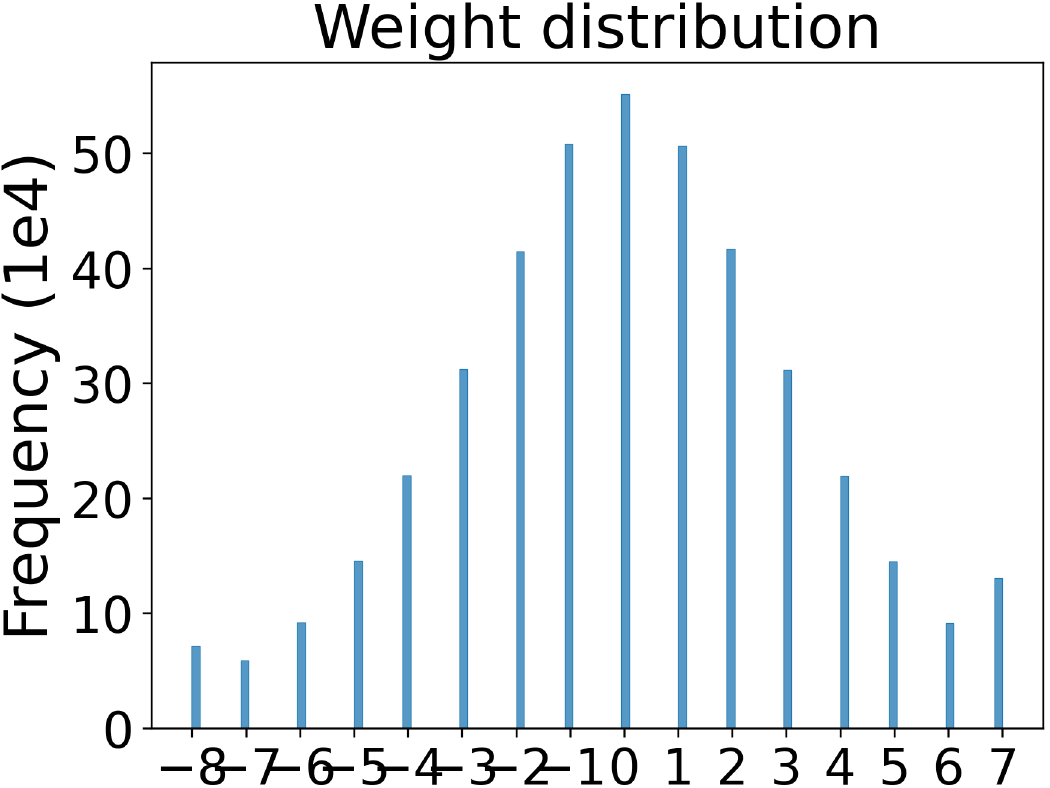}
    \label{fig:sub2}}
  \hfil 
  \subfloat[]{\includegraphics[width=0.34\textwidth]{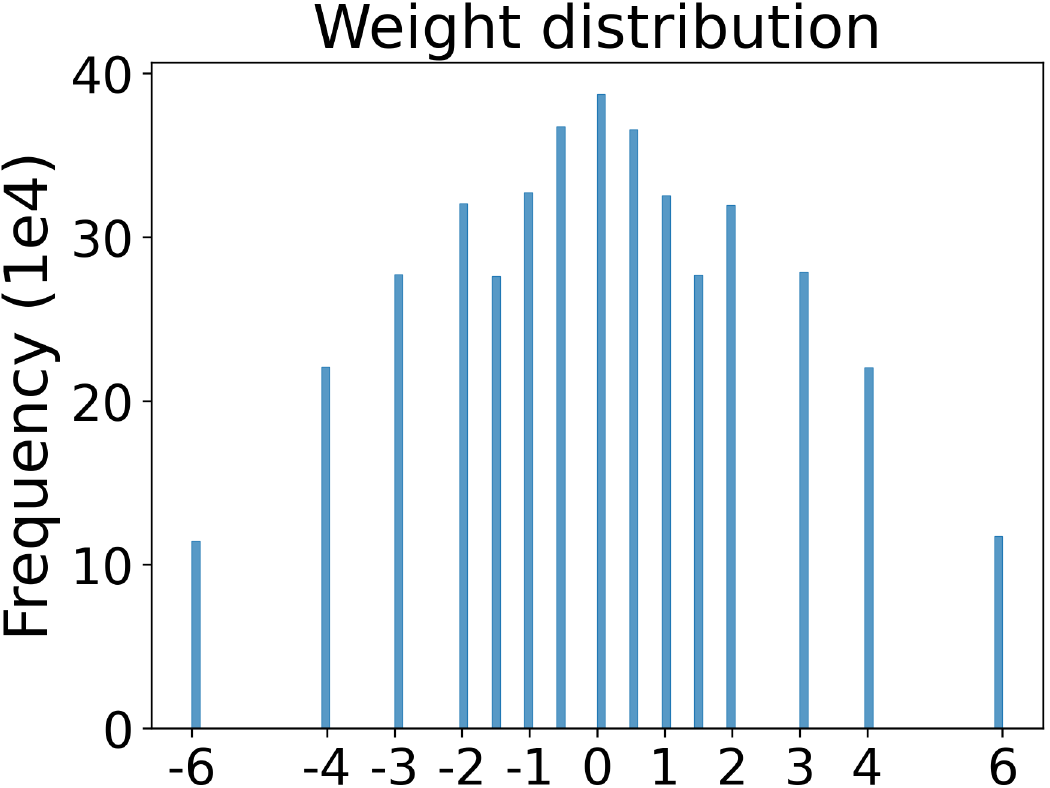}
    \label{fig:sub3}}
  \hfil
    \subfloat[]{\includegraphics[width=0.34\textwidth]{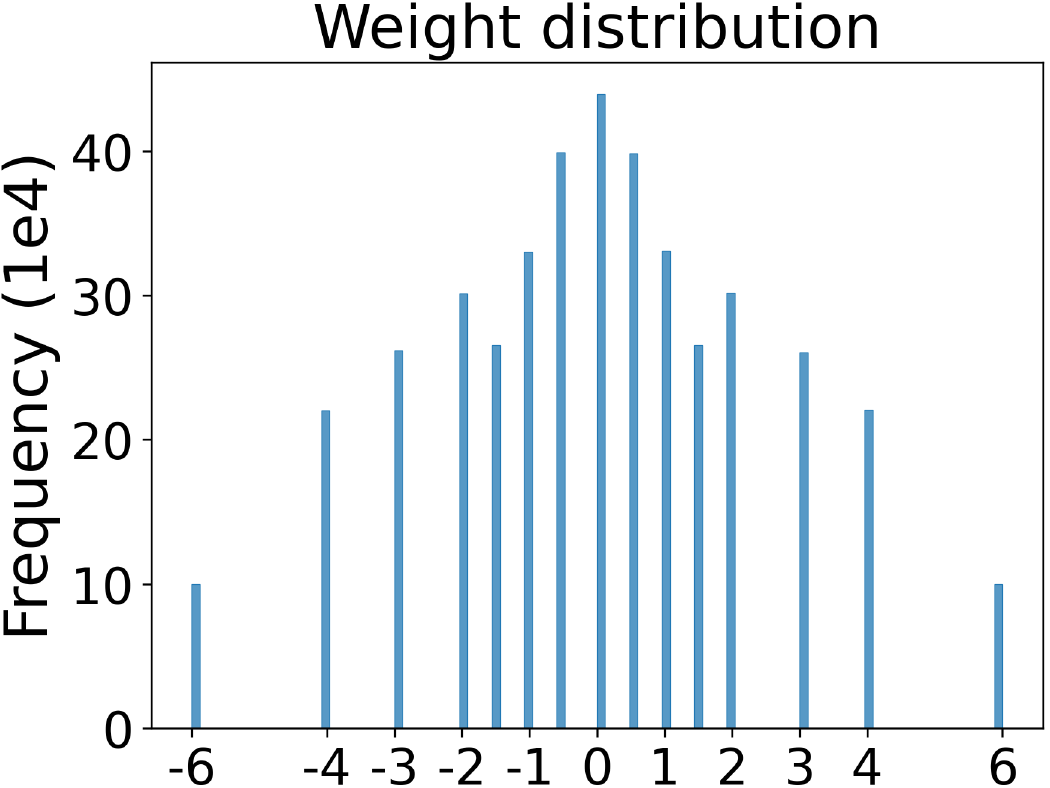}
    \label{fig:sub4}}
  \caption{Weight distribution of a q\_proj layer of LlaMA-3.2-1B. (a): Weight distribution in full-precision(BF16). (b): Weight distribution after INT4 quantization (group size=32), compared with MXFP4. (c): Weight distribution after MXFP4 quantization. (d): Weight distribution after NVFP4 quantization.}
  \label{fig:three_images}
\end{figure}

FP4 can better handle long-tail distributions compared to INT4, as its values are non-uniformly distributed. In LLMs, both weights and activations approximately follow normal distributions, with activations particularly exhibiting long-tail characteristics. Fig. \ref{fig:three_images} shows the weight distribution under BF16, INT4, MXFP4 and NVFP4. A clear observation is that under INT4 quantization, values with larger absolute magnitudes represent fewer data points, whereas MXFP4 more fully utilizes the representational capacity of 4-bit precision. Therefore, FP4 is more suitable than INT4 as a quantization data type.

LLM-FP4 \cite{liu2023llm} studies FP4 precision quantization for LLMs, including searching for the optimal exponent bias and maximal quantization value. It also proposes a pre-shifted exponent bias technique to address the issue of high inter-channel variance. FP4-Quantization \cite{wang2024fp4} improves the FP4 data representation while introducing a method called LREC (low-rank error calibration) to quantize weights into FP4.

\section{Extensive Evaluations}
\textbf{Evaluation Scope.}
Based on the previous review and the defined experiment scope, we conducted extensive evaluations to systematically investigate the effects of Post-Training Quantization (PTQ) on large language models. Our analysis is organized into three main parts. In Section \ref{sec:3.1}, we evaluate several popular methods within the two-step quantization framework, providing a baseline for comparison. Section \ref{sec:3.2} performs an ablation study across different quantization settings, examining how these choices influence model performance. Section \ref{sec:3.3} applies the evaluated methods to FP4 and its variants, further exploring the potential and limitations of FP4 quantization.

\textbf{Quantization settings.}
We conduct experiments using W4A4 precision, comparing the results with the original precision (BF16 or FP16). In the main results, we use per-token or per-group (group size=128) asymmetric quantization for activations. For weight quantization, we use per-channel or per-group (group size=128) symmetric quantization, where we extract the clipping ratio using a linear search over the squared error.
Experiments are conducted on NVIDIA A6000/H200 GPUs with CUDA 12.4 (except latency experiments). For calibration, we use 128 samples from WikiText-2 (sequence length 2048) for GPTQ and low-rank compensation, and 1,000 samples for training learnable rotation/scaling matrices, optimized with RiemannAdam. Calibrated scaling follows SmoothQuant~\cite{xiao2023smoothquant} with $\alpha=0.5$. The low-rank compensation rank is $k=32$, stored in FP16, following the common setup in ZeroQuant-v2~\cite{yao2023zeroquant} and LQER~\cite{zhang2024lqer}. For rotation and scaling optimization, we adopt each method's official training procedure. For clipping ratio in symmetric weight quantization, we use linear search over squared error. Random seeds used in experiments are $\{0,1,2,3,4,42\}$. See Appendix for code sources.

\textbf{Models and tasks.}
We conduct experiments on LLaMA-3.2 (1B, 3B), LLaMA-3.1 (8B, 70B), and Qwen2.5-3B. Additionally, we perform experiments on the MoE model Qwen1.5-MoE-A2.7B to further extend our evaluation. We evaluate their perplexity on the WikiText-2 dataset \cite{merity2016pointer} and assess their accuracy on several zero-shot evaluation tasks, including commonsense reasoning, mathematics, sentiment analysis, and question answering. Specifically, we utilize the following datasets for evaluation: MMLU \cite{hendryckstest2021}, SST2 \cite{socher-etal-2013-recursive}, MathQA \cite{amini2019mathqa}, ARC‑C \cite{boratko2018systematic}, HellaSwag \cite{zellers2019hellaswag}, WinoGrande \cite{sakaguchi2019winogrande}, BoolQ \cite{clark2019boolq}, LAMBADA \cite{radford2019language}, and PIQA \cite{bisk2020piqa}.
\begin{table}[h]
\centering
\renewcommand{\arraystretch}{0.8}
\begin{tabular}{cc cc cc}
\toprule
\multicolumn{2}{c}{Evaluate Methods} & \multicolumn{2}{c}{Channel}
 & \multicolumn{2}{c}{Group-128} \\

PT & QEM  &\makecell{Wiki2\\PPL($\downarrow$) }& \makecell{0‑shot\textsuperscript{9}\\ Avg.($\uparrow$) }
& \makecell{Wiki2\\PPL($\downarrow$) }& \makecell{0‑shot\textsuperscript{9}\\ Avg.($\uparrow$) }\\ 
\midrule 
\multicolumn{2}{c }{BF16} & 9.76 & 54.78 & 9.76 & 54.78 \\
\midrule 
Opt. Rot& RTN	&	17.83	&	45.24	&	14.84	&	47.51	\\
Opt. Rot& Low-rank	&	17.16	&	45.47	&	14.50	&	47.04	\\
Opt. Rot& GPTQ	&	12.95	&	47.79	&	11.65	&	49.38	\\
Opt. Rot& GPTQ+Low-rank	&	12.81	&	48.35	&	11.61	&	50.49	\\


\midrule

Opt. Rot+Sca& RTN	&	14.88	&	46.41	&	13.3	&	49.22	\\
Opt. Rot+Sca& Low-rank	&	14.58	&	46.76	&	13.07	&	49.95	\\
Opt. Rot+Sca& GPTQ	&	11.80	&	48.89	&	10.89	&	50.63	\\
Opt. Rot+Sca& GPTQ+Low-rank	&	11.73	&	49.06	&	10.87	&	51.83	\\

\bottomrule 
\end{tabular}
\caption{Evaluation of quantization method combinations, in which rotation or scaling is optimized, in terms of Perplexity on WikiText2 and averaged accuracy on nine Zero-Shot tasks across per-channel and per-group settings of LlaMA-3.2-1B. PT and QEM represent pre-quantization transformation and quantization error mitigation, respectively. Rot and Sca denote Rotation and Scaling, respectively, and Opt. indicates optimized.}\label{tab:experiment1}
\end{table}

\begin{figure}[ht]
\centering
\includegraphics[width=0.8\textwidth]{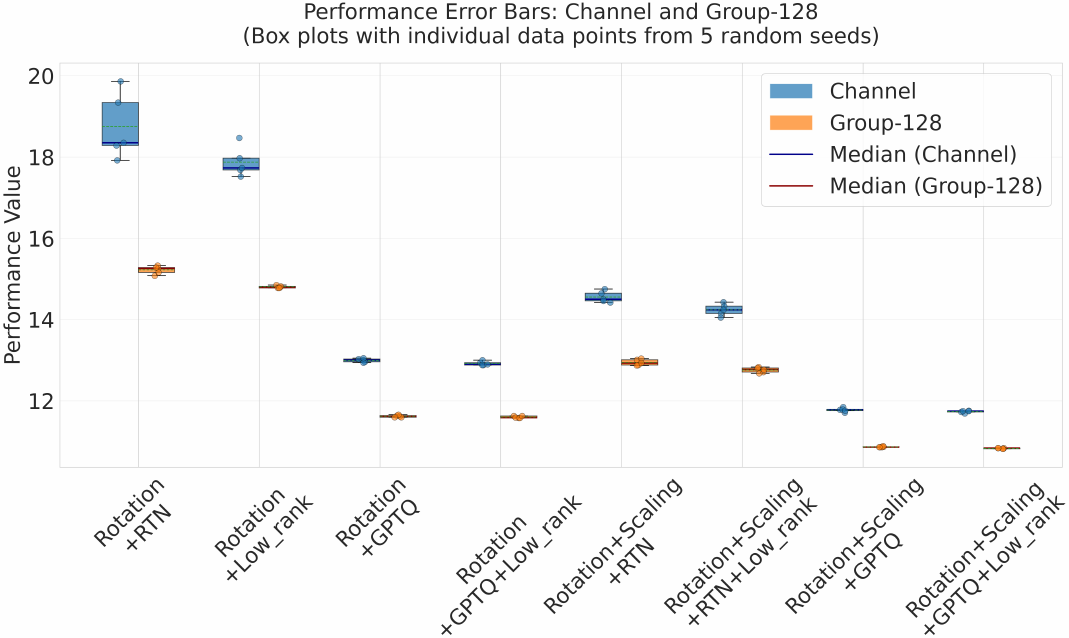}
  \caption{Error bars showing the statistical robustness of the methods in Table \ref{tab:experiment1}, computed over five runs with different random seeds.}
  \label{fig:error-bar}
\end{figure}

\begin{table}[t]
\centering
\renewcommand{\arraystretch}{0.8}
\begin{tabular}{cccc}
\toprule
\multicolumn{2}{c}{Evaluate Methods} & Channel & Group-128 \\
PT & QEM  &\multicolumn{2}{c}{Wiki2 PPL($\downarrow$) } \\ 
\midrule
 \multicolumn{2}{c}{BF16} & \multicolumn{2}{c}{9.76} \\ 
\midrule
Opt. Rot & RTN & 18.75 $\pm$ 0.814 & 15.22 $\pm$ 0.101 \\
Opt. Rot & Low-rank & 17.87 $\pm$ 0.370 & 14.80 $\pm$ 0.030 \\
Opt. Rot & GPTQ & 13.00 $\pm$ 0.046 & 11.62 $\pm$ 0.026 \\
Opt. Rot & GPTQ+Low-rank & 12.92 $\pm$ 0.051 & 11.60 $\pm$ 0.026 \\
\midrule
Opt. Rot+Sca & RTN & 14.56 $\pm$ 0.139 & 12.95 $\pm$ 0.076 \\
Opt. Rot+Sca & RTN+Low-rank & 14.24 $\pm$ 0.149 & 12.76 $\pm$ 0.064 \\
Opt. Rot+Sca & GPTQ & 11.78 $\pm$ 0.051 & 10.87 $\pm$ 0.013 \\
Opt. Rot+Sca & GPTQ+Low-rank & 11.74 $\pm$ 0.028 & 10.83 $\pm$ 0.011 \\
\bottomrule
\end{tabular}
\caption{Statistical robustness analysis of different quantization pipelines for LlaMA-3.2-1B under 5 random seeds. Results are reported as $\text{Mean} \pm \text{Standard Deviation (SD)}$ in terms of perplexity (PPL). The lower, the better.}\label{tab:statistical_robustness}
\end{table}
\begin{table}[!t]
\centering
\renewcommand{\arraystretch}{0.8}
\begin{tabular}{cc cccc}
\toprule
\multicolumn{2}{c }{Evaluate Methods} & \multicolumn{2}{c }{Channel}
 & \multicolumn{2}{c}{Group-128} \\

PT & QEM  &\makecell{Wiki2\\PPL($\downarrow$) }& \makecell{0‑shot\textsuperscript{9}\\ Avg.($\uparrow$) }
& \makecell{Wiki2\\PPL($\downarrow$) }& \makecell{0‑shot\textsuperscript{9}\\ Avg.($\uparrow$) }\\ 
\midrule 
\multicolumn{2}{c }{BF16} & 9.76 & 54.78 & 9.76 & 54.78 \\
\midrule

Rot& RTN & 20.31 & 43.26 & 15.18 & 46.62 \\
Rot& Low‑rank & 19.84 & 43.09 & 14.83 & 46.87 \\
Rot & GPTQ & 13.31 & 48.02 & 11.96 & 50.05 \\
Rot & GPTQ+Low‑rank & 13.28 & 48.37 & 11.98 & 50.48 \\
\midrule
Rot+Sca& 	RTN   & 21.15& 	40.89	& 15.50& 	44.12\\
Rot+Sca	& RTN+Low‑rank	& 20.93& 	41.63 &   15.07	& 44.93\\
Rot+Sca& 	GPTQ	& 13.80& 	45.22 &  12.20	& 47.54\\
Rot+Sca	& GPTQ+Low‑rank  & 13.76	& 45.39 & 	12.08	& 47.77\\

\bottomrule 
\end{tabular}
\caption{Evaluation of quantization method combinations, in which rotation or scaling is non-optimized, in terms of Perplexity on WikiText2 and averaged accuracy on nine Zero-Shot tasks across per-channel and per-group settings of LlaMA-3.2-1B.}\label{tab:experiment2}
\end{table}
\subsection{Methods evaluation under the two-step quantization framework}
\label{sec:3.1}

\begin{table}[ht]
\centering
\setlength{\tabcolsep}{3pt}
\renewcommand{\arraystretch}{0.8}
\begin{tabular}{l cccc cccc}
\toprule
\multirow{3}*{Evaluate Methods}  & \multicolumn{4}{c}{Channel Wiki2 PPL ($\downarrow$)} & \multicolumn{4}{c}{Group-128 Wiki2 PPL ($\downarrow$)} \\
\cmidrule(lr){2-5} \cmidrule(lr){6-9}
& \multicolumn{2}{c}{w/o Opt.} & \multicolumn{2}{c}{w/ Opt.} & \multicolumn{2}{c}{w/o Opt.} & \multicolumn{2}{c}{w/ Opt.} \\
\cmidrule(lr){2-3} \cmidrule(lr){4-5} \cmidrule(lr){6-7} \cmidrule(lr){8-9}
 & GPTQ & G+LR & GPTQ & G+LR & GPTQ & G+LR & GPTQ & G+LR \\
\midrule
BF16 & \multicolumn{4}{c}{9.76} & \multicolumn{4}{c}{9.76} \\
\midrule
Rot             & 13.31 & 13.28 & 12.95 & 12.81 & 11.96 & 11.98 & 11.65 & 11.61 \\
Rot + Shi       & 13.71 & 13.57 & 12.69 & 12.58 & 12.06 & 11.96 & 11.57 & 11.51 \\
Rot + Sca       & 13.80 & 13.76 & 11.80 & 11.73 & 12.20 & 12.08 & 10.89 & 10.87 \\
Rot + Sca + Shi & 14.12 & 13.98 & 11.66 & 11.65 & 12.27 & 12.20 & 10.85 & 10.80 \\
\bottomrule
\end{tabular}
\caption{Evaluation of pre-quantization transformations with (w/) and without (w/o) optimization under GPTQ and GPTQ (G) +Low-rank (LR) settings on Llama-3.2-1B.}\label{tab:eval_transformations}
\end{table}
\begin{table}[!t]
\centering
\renewcommand{\arraystretch}{0.8}
\begin{tabular}{cc
p{1.0cm} p{1.0cm}
p{1.0cm} p{1.0cm}
p{1.0cm} p{1.0cm}}
\toprule
\multicolumn{2}{c}{\multirow{2}*{Metric}} 
& \multicolumn{2}{c}{LLaMA-3.2-1B} 
& \multicolumn{2}{c}{LLaMA-3.2-3B} 
& \multicolumn{2}{c}{Qwen2.5-3B} \\
\cmidrule(lr){3-4} \cmidrule(lr){5-6} \cmidrule(lr){7-8}
&& FP16 & W4A4 & FP16 & W4A4 & FP16 & W4A4 \\
\midrule
\multicolumn{2}{c}{WikiText2 PPL($\downarrow$)} 
& 9.76 & 11.73 & 7.82 & 8.90 & 8.03 & 9.22 \\
\midrule
 General & MMLU & 36.76 & 29.21 & 54.06 & 46.70 & 65.13 & 56.32 \\

 Sentiment & SST2 & 66.86 & 55.28 & 73.62 & 68.58 & 90.37 & 69.15 \\

 Math & MathQA & 28.94 & 25.43 & 34.47 & 30.55 & 37.15 & 32.50 \\

 \multirow{3}*{Reasoning} & PiQA & 74.65 & 69.15 & 77.48 & 73.99 & 78.73 & 74.16 \\
& HellaS & 63.67 & 57.93 & 73.58 & 69.54 & 73.55 & 67.52 \\
& WinoG & 60.38 & 56.43 & 69.46 & 65.43 & 68.75 & 62.12 \\

 \multirow{3}*{QA} & BoolQ & 63.67 & 59.02 & 72.97 & 68.75 & 77.58 & 65.41 \\
& Lambada & 62.10 & 56.78 & 70.08 & 65.32 & 66.80 & 57.77 \\
& Arc-C & 36.01 & 32.34 & 46.08 & 41.55 & 47.35 & 42.83 \\
 \multicolumn{2}{c}{Average} 
& 54.78 & 49.06 & 63.53 & 58.93 & 67.27 & 58.64 \\
\midrule
\multicolumn{2}{c}{\multirow{2}*{Metric}} 
& \multicolumn{2}{c}{Qwen-MoE-A2.7B} 
& \multicolumn{2}{c}{LLaMA-3.1-8B} 
& \multicolumn{2}{c}{LLaMA-3.1-70B} \\
\cmidrule(lr){3-4} \cmidrule(lr){5-6} \cmidrule(lr){7-8}
&& FP16 & W4A4 & FP16 & W4A4 & FP16 & W4A4 \\
\midrule
\multicolumn{2}{c}{WikiText2 PPL($\downarrow$)} 
& 7.22 & 8.88 & 6.24 & 7.23 & 2.81 & 4.41 \\
\midrule
General & MMLU  & 60.86 & 51.38 & 63.32 & 56.98 & 75.26 & 70.57 \\
Sentiment & SST2  & 67.89 & 60.32 & 77.18 & 59.52 & 82.11 & 75.69 \\
Math & MathQA  & 35.01 & 29.21 & 39.50 & 34.91 & 51.66 & 45.36 \\
 \multirow{3}*{Reasoning} & PiQA  & 80.69 & 76.39 & 81.18 & 78.51 & 84.22 & 81.12 \\
& HellaS &  77.30 & 72.08 & 78.92 & 76.05 & 85.03 & 83.37 \\
& WinoG  & 69.38 & 62.59 & 73.40 & 68.90 & 79.79 & 76.24 \\
\multirow{3}*{QA}  & BoolQ  & 79.60 & 73.98 & 82.05 & 78.75 & 85.50 & 83.39 \\
& Lambada  & 71.38 & 61.69 & 75.33 & 72.48 & 78.75 & 76.75 \\
& Arc-C  & 44.37 & 39.76 & 53.75 & 48.29 & 64.85 & 59.98 \\
\multicolumn{2}{c}{Average} 
& 65.16 & 58.60 & 69.40 & 63.82 & 76.35 & 72.50 \\
\bottomrule
\end{tabular}
\caption{Evaluation of performance under different models. }
\label{tab:experiment3}
\end{table}

We select three of the most widely used pre-quantization transformation techniques, shifting, scaling and rotation, and combine them with two quantization error mitigation methods: GPTQ and low-rank compensation. A series of experiments is conducted to evaluate these combinations. Table \ref{tab:experiment1} presents the results of using optimized scaling and rotation, illustrating the performance of different method combinations. Table \ref{tab:experiment2} presents the results obtained with randomly initialized rotations and calibrated scaling. Table \ref{tab:eval_transformations} presents the results obtained with three transformations combinations.



To rigorously demonstrate that the observed performance trends are reliable and not artifacts of stochastic variation, we conduct extensive experiments across five independent random seeds ($0$ to $4$). The comprehensive statistics ($\text{Mean} \pm \text{SD}$) are reported in Table~\ref{tab:statistical_robustness}, with the corresponding variance distributions visualized via the error bars in Fig.~\ref{fig:error-bar}. We observe that as advanced optimization components are progressively integrated, the performance variance across random seeds diminishes drastically. For instance, under the channel-wise setting, the standard deviation drops sharply from $0.814$ (Rotation+RTN) to an exceptionally low $0.028$ (Rotation+Scaline+GPTQ+Low-rank). 

In Table \ref{tab:experiment3}, we further conduct experiments on different models. The experiments of LLaMA-3.2-1B, LLaMA-3.2-3B, LLaMA-3.1-8B, LLaMA-3.1-70B, and Qwen2.5-3B use the optimized rotation and scaling, and combine them with GPTQ and an additional low-rank compensation. 
For the MoE model Qwen1.5-MoE-A2.7B, given that the key advantage of MoE models lies in the fact that the actual computational cost scales only with the number of activated experts and performs full optimization of rotation and scaling parameters and GPTQ would be computationally expensive in quantization process compared to models with comparable computational budgets, we opt for more resource-efficient strategies which only use random rotation and low-rank compensation during PTQ.

In the following subsections, we provide a comprehensive and systematic analysis of the experimental results from multiple perspectives.
\subsubsection{Pre-quantization transformation}

\textbf{Under optimization, using more optimized transformations yields better performance.} As shown in Table \ref{tab:experiment1}  and  \ref{tab:eval_transformations}, applying optimized scaling or shifting together with optimized rotation always yields better performance. This is mainly because, after optimization, scaling and shifting can effectively fine-tune the inter-channel magnitude distribution of the rotated tensor, further improving the results and achieving better spatial utilization. However, shifting yields marginal performance add compared with applying optimized rotation and scaling. Using only rotation also achieves significantly better performance compared with random rotation, which is consistent with the observation in SpinQuant \cite{liu2024spinquant} that optimization helps eliminate the high variance introduced by random rotations.

\textbf{Without optimization, it is better to use random rotation only.} When the model is large, performing optimization is often impractical; therefore, we can only rely on transformation parameters obtained through calibration. From Table \ref{tab:experiment2} and \ref{tab:eval_transformations}, we can see that in the absence of optimization, rotation alone performs better than rotation combined with shifting or scaling.
We believe this is because when the shifting or scaling is applied after rotation, we hope it can fine-tune the rotated matrix. However, the calibration method becomes inaccurate in this case because outliers are averaged across channels, and each channel's shifting or scaling parameter cannot be simply calculated using activation distributions.

\subsubsection{Quantization error mitigation}
\textbf{Comparison between self-compensation and low-rank compensation.} Between the two quantization error mitigation methods, low-rank compensation is often inferior to GPTQ, except in per-channel scaling without transformation. Under optimized rotation and scaling, GPTQ alone achieves a perplexity of 11.8, compared to 14.58 for low-rank compensation. This is likely because our low-rank method performs SVD on the weight quantization error and thus cannot approximate the loss error as effectively as GPTQ. Loss-aware low-rank compensation may achieve better results but requires more complex computation, which we leave for future study.

\textbf{Combination of self-compensation and low-rank compensation}. 

When both methods are deployed in tandem, in most cases, the performance improves further, decreasing the perplexity from $11.78$ to $11.74$ in the per-channel setting, and from $10.87$ to $10.83$ in the per-group setting, thereby further narrowing the gap to the full-precision model. We attribute this enhancement to the fact that performing low-rank compensation on the residual weight quantization error after GPTQ provides an orthogonal layer of error mitigation, leading to superior results. 

Notably, although the absolute numerical margins here appear relatively narrow, a paired $t$-test on PPL results confirms that this collaborative improvement is highly reproducible and statistically significant, yielding a $p$-value of $0.0224$ ($p < 0.05$) for the per-channel configuration. This firmly rejects the null hypothesis and validates that the joint mechanism brings a reliable, non-trivial boost to quantization quality.

\subsubsection{Relationship between the two-step}
\textbf{The performance contributions of these two steps are additive and mutually complementary.} Our experiments reveal that the best performance is achieved when both optimized scaling and rotation are applied in conjunction with GPTQ and the low-rank compensation. This configuration yields a perplexity of 11.73 in per-channel quantization for the LLaMA-3.2-1B model, which is close to BF16 of 9.76.

\subsection{Evaluation of quantization settings}
\label{sec:3.2} 
\subsubsection{Symmetric vs. asymmetric}
\begin{table}[ht]
\centering
\renewcommand{\arraystretch}{0.8}
\begin{tabular}{cccccc} 
\toprule 
\multirow{2}*{Granularity}& \multirow{2}*{Metric} & \multicolumn{4}{c}{Rotation+Scaling+GPTQ}\\
 & &{Sym}& {W-Asym}& {A-Asym}& {Asym}\\
\midrule
\multirow{2}*{Channel}&Wiki2-PPL($\downarrow$)&12.73	&	12.7	&	11.8	&	11.7		\\
&Arc-C Score(\%)& 31	&	31.7	&	33.3	&	33.7
\\
\midrule
\multirow{2}*{Granularity}& \multirow{2}*{Metric} & \multicolumn{4}{c}{Rotation+Scaling+GPTQ+Low-rank}
\\
 & &{Sym}& {W-Asym}& {A-Asym}& {Asym}\\
\midrule
\multirow{2}*{Channel}&Wiki2-PPL($\downarrow$)	&		12.63	&	12.65	&	11.73	&	11.64	\\
&Arc-C Score(\%)&	31.9	&	29.9	&	32.3	&	32.7
\\
\bottomrule 
\end{tabular}
\caption{Comparison of symmetric and asymmetric quantization in terms of Perplexity on WikiText2 and Arc \cite{boratko2018systematic} accuracy of LlaMA-3.2-1B}\label{tab:sym}
\end{table}


Table \ref{tab:sym} compares symmetric and asymmetric quantization for weights and activations. Asymmetric quantization substantially improves activation quantization but provides only minor gains for weights, likely because activations tend to exhibit more asymmetric distributions while weights are generally more symmetric. Therefore, using symmetric quantization for weights and asymmetric quantization for activations provides a favorable trade-off, significantly outperforming fully symmetric quantization while reducing the dequantization overhead of fully asymmetric quantization.

\subsubsection{Granularity}
\begin{figure}[t]
\centering
\includegraphics[width=0.7\linewidth]{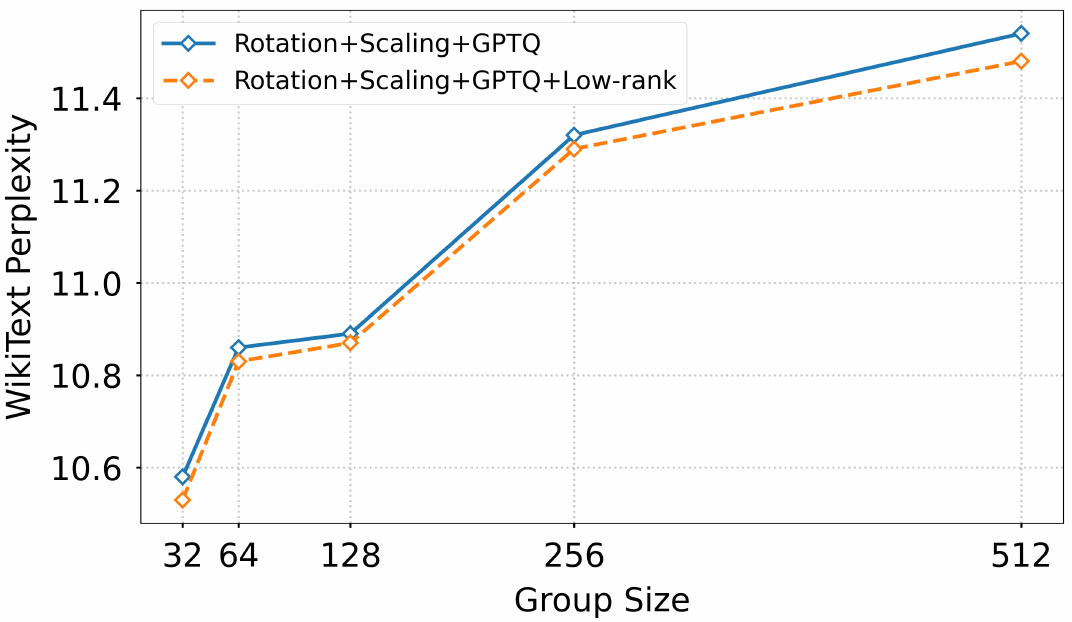}
\caption{Comparison of different group sizes of quantization in terms of Perplexity on WikiText2 and averaged accuracy on Zero-Shot tasks of LlaMA-3.2-1B}\label{fig: granularity}
\label{fig:sym}
\end{figure}

As illustrated in Fig. \ref{fig: granularity}, we conduct a comprehensive comparison of per-group quantization performance across varying group sizes, measured by perplexity on WikiText2. The experimental results reveal that finer quantization granularity yields superior model performance (lower perplexity), albeit at the expense of elevated storage requirements for quantization parameters. This trade-off stems primarily from the scaling factors, whose storage overhead scales inversely with group size. Quantitative analysis of these additional bit costs is presented in Table \ref{tab: granularity}, which compares the average bit overhead introduced by scaling factors under symmetric weight quantization.

\begin{table}[H]
\centering
\renewcommand{\arraystretch}{0.8}
\begin{tabular}{cccccc} 
\toprule 
Granularity& group-32& group-64&group-128&group-256&group-512\\
\midrule
{\makecell{Average Extra bits}}& 1 & 0.5&0.25&0.125&0.0625\\
\bottomrule 
\end{tabular}
\caption{Average extra bits introduced by FP32 scaling factors of different granularities}
\label{tab: granularity}
\end{table}

\subsection{Extending Experiments in FP4}
\label{sec:3.3}
\subsubsection{FP4 vs. INT4}
\label{sec:fp4vsint4}
Extending our experiments to the FP4 format, we compare several variants of FP4, including MXFP4, NVFP4, and a naive FP4 implementation with varying granularities and scaling factor formats. For per-channel quantization, we employ asymmetric quantization for activations and symmetric quantization for weights. To ensure a fair comparison with MXFP4 and NVFP4, all other experiments using per-group quantization adopt symmetric quantization for both activations and weights.



\begin{table}[ht]
\centering
\renewcommand{\arraystretch}{0.8}
\begin{tabular}{ccccc}
\toprule
\multirow{2}{*}{Data Format}
& INT4 & FP4 & INT4 & FP4 \\
& channel, FP32 & channel, FP32 & g16, FP32 & g16, FP32 \\
\midrule
RTN             & 256.97 & 135.68 & 11.75 & 11.34 \\
GPTQ            & 161.04 & 101.36 & 11.47 & 10.87 \\
Low-rank        & 179.87 & 132.14 & 11.69 & 11.32 \\
GPTQ+Low-rank   & 139.31 & 96.33  & 10.92 & 10.76 \\
\midrule
\multirow{2}{*}{Data Format}
& MXFP4 & FP4 & FP4 & NVFP4 \\
& g32, E8M0 & g32, E4M3 & g16, E8M0 & g16, E4M3 \\
\midrule
RTN             & 15.17 & 12.04 & 14.52 & 11.54 \\
GPTQ            & 13.24 & 11.36 & 12.77 & 10.98 \\
Low-rank        & 15.56 & 11.98 & 14.67 & 11.48 \\
GPTQ+Low-rank   & 12.69 & 11.18 & 12.25 & 10.83 \\
\bottomrule
\end{tabular}
\caption{Evaluation of INT4 and FP4 quantization without pre-quantization transformation on LLaMA-3.2-1B in terms of WikiText2 PPL. Channel denotes per-channel quantization, while g16/g32 denotes per-group quantization with group sizes of 16/32. FP32, E8M0, and E4M3 refer to the format of the scaling factor. For E4M3, the scaling strategy follows NVFP4’s approach, employing a two-level scaling scheme.}
\label{tab:fp4-1}
\end{table}

First, we apply no extra pre-quantization transformation, conducting experiments using only quantization error mitigation methods.

\textbf{FP4 performs better than INT4 without pre-quantization transformation in per-channel setting.}
Table \ref{tab:fp4-1} demonstrates that FP4 (per-channel) shows much better performance than INT4 (per-channel). We believe this is because, in the absence of the pre-quantization transformation, FP4 better aligns with the data distribution, thus achieving superior results.

\textbf{For finer granularity scenarios, FP4 and INT4 show very small differences.} Table \ref{tab:fp4-1} demonstrates that FP4 and INT4 show minimal differences at finer granularities, with FP4 performing only slightly better than INT4. We attribute this to the fact that fine-grained quantization helps mitigate the impact of outliers. Consequently, the advantage of FP4 in handling outliers is reduced under finer granularity, resulting in only a marginal performance improvement over INT4.

\textbf{Compared to MXFP4, NVFP4 achieves better performance due to its smaller group size and more precise scaling factor format.} Both the format of the scaling factor and the size of the group significantly impact performance. With the E4M3 variant of FP8, NVFP4 encodes scaling factors with fractional precision, whereas MXFP4 uses E8M0, restricting scales to powers of two and causing coarser quantization. To compensate for the limited range of E4M3, NVFP4 further employs two-level scaling with a per-tensor FP32 scale. These designs reduce the risk of accuracy degradation, consistent with NVIDIA's findings~\cite{nvidianvfp4}.

\begin{table}[ht]
\centering
\renewcommand{\arraystretch}{0.8}
\begin{tabular}{cccc@{\hspace{3pt}}c@{\hspace{3pt}}c}
\toprule 
\multicolumn{2}{c}{Evaluate Methods}& \multicolumn{4}{c}{Wiki2‑PPL($\downarrow$)}\\

\makecell{PT}&\makecell{QEM}&\makecell{INT4\\C, FP32}&\makecell{FP4\\C, FP32}&\makecell{ MXFP4\\g32, E8M0} & \makecell{NVFP4\\g16, E4M3}\\
\midrule 

Sca&GPTQ         &134.85 &  89.22   &13.24 &10.97\\
Sca&Low-rank     & 165.4 &  104.55   &14.91 &11.36\\
Sca&GPTQ+Low-rank& 124.56&  114.96   &12.58  & 10.81\\
Opt. Rot+Sca & GPTQ& 11.8& 12.13  &12.37 &11.04\\
Opt. Rot+Sca& Low-rank& 14.58 &15.15  &15.43 &12.57\\
Opt. Rot+Sca & GPTQ+Low-rank&11.73 &12.13 & 12.29& 11.01\\
\bottomrule 
\end{tabular}
\caption{Evaluation of INT4, MXFP4, and NVFP4 using both pre-quantization transformation and quantization error mitigation techniques, in terms of Perplexity on WikiText2 of LlaMA-3.2-1B.}
\label{tab:fp4-2}
\end{table}
Then, we apply pre-quantization transformation together with quantization error mitigation on FP4 and main FP4 variants to compare with INT4, and the results are shown in Table \ref{tab:fp4-2}.

\textbf{Rotation-based methods yield smaller improvements for MXFP4 and NVFP4 than for INT4 (per-channel).} As argued in AMXFP4 \cite{lee2024amxfp4}, grouping activation tensors into small micro-scaled units mitigates outliers like rotation. For MXFP4 and NVFP4 formats, the group size is extremely small, so data rotation contributes little and may even lead to degradation.
  
\textbf{FP4 does not show its advantage under rotation.} Although MXFP4 and NVFP4 perform better than INT4 (per-channel), it is observed that under the same channel-wise setting, INT4 performs better than FP4 after rotation, while when applying only scaling, FP4 still outperforms INT4. We believe this is because rotation can largely eliminate outliers, to the extent that the advantage of FP4 over INT4 when dealing with outliers diminishes. Consequently, INT4 demonstrates stronger performance in handling more uniform distributions.

\subsubsection{Exploration of FP4 scaling factor}
\begin{table}[h]
\centering 
\renewcommand{\arraystretch}{0.8}
\begin{tabular}{cccccc} 
\toprule 
\multirow{4}*{\makecell{Quantization \\error\\ mitigation}} &\multicolumn{5}{c}{Data Format}\\
\cmidrule(lr){2-6}

&\makecell{NVFP4\\W:E4M3\\A:E4M3} 
&\makecell{FP4\\W:E2M1\\A:E4M3} 
&\makecell{FP4\\W:E2M1\\  A:E2M1} 
&\makecell{FP4\\W:INT4\\A:E4M3} 
&\makecell{FP4\\W:INT4\\A:INT4} \\
\midrule

RTN  	        & 11.54  & 14.10  & 4130 & 14.66 & 4777 \\
GPTQ	        & 10.98  & 13.38  & 3618 & 17.96 & 3803 \\
Low-rank 	    & 11.48  & 13.75  & 4057 & 14.43 & 5356 \\
GPTQ+Low-rank  & 10.83  & 11.38  & 3650 & 11.59 & 4124 \\
\bottomrule 
\end{tabular}
\caption{Evaluation of FP4 format quantization using low-bit scaling factors and group size 16, in terms of WikiText2 PPL on LlaMA-3.2-1B. E4M3, E2M1, and INT4 denote the scaling-factor formats, and W/A denote weights/activations. All of these follow NVFP4's two-level scaling scheme.}\label{tab:fp4-3}
\end{table}
In the analysis of Table \ref{tab:fp4-1} in Section \ref{sec:fp4vsint4}, we observed that the scaling factor has a significant impact, and we conduct further exploratory research on this aspect. We further reduced the bit-width of the scaling factor to 4 bits to evaluate its effectiveness under such a low-bit setting. As shown in Table \ref{tab:fp4-3}, applying 4-bit scaling factors to both weights and activations leads to very poor performance. When applied only to weights, the performance remains moderate but is worse compared to NVFP4.

\subsection{Effective Evaluation}
\subsubsection{Inference Latency}
To further validate the practical efficiency gains, we measured inference latency under different batch sizes on LLaMA-3.1-8B. For INT4 compared with BF16, experiments are conducted under CUDA 12.4 / PyTorch 2.6 on RTX 4090. For NVFP4 compared with BF16, experiments are conducted under CUDA 13.0 / PyTorch 2.12 on RTX 5090. Following standard evaluation practices for quantization kernels, we measured the latency at the block level with a prefill sequence length of 2048 and 256 decoding steps. All measurements were conducted over 30 benchmark iterations after 10 warm-up runs. We report the prefill latency and the average per-step decoding latency.
\begin{table}[ht]
\centering
\renewcommand{\arraystretch}{0.8}
\setlength{\tabcolsep}{1.5pt} 
\begin{tabular}{@{} l l l c c c c @{}}
\toprule
\textbf{Stage} & \textbf{GPU} & \textbf{Config} & \textbf{B=1} & \textbf{B=4} & \textbf{B=16} & \textbf{B=32} \\
\midrule
\multirow{6}{*}{\shortstack[l]{\textbf{Prefill}\\(ms)}} 
 & \multirow{3}{*}{4090} & BF16 & 8.75 & 35.10 & 133.71 & 266.75 \\
 & & INT4 & 3.96 (2.21$\times$) & 15.60 (2.25$\times$) & 68.36 (1.96$\times$) & 136.22 (1.96$\times$) \\
 & & INT4-Rot & 4.38 (2.00$\times$) & 17.21 (2.04$\times$) & 74.59 (1.79$\times$) & 148.59 (1.80$\times$) \\
 \cmidrule(lr){2-7}
 & \multirow{2}{*}{5090} & BF16 & 4.98 & 24.12 & 99.99 & 166.11 \\
 & & NVFP4 & 2.45 (2.03$\times$) & 10.46 (2.31$\times$) & 49.98 (2.00$\times$) & 99.98 (1.66$\times$) \\
\midrule
\multirow{6}{*}{\shortstack[l]{\textbf{Decode}\\(ms/step)}} 
 & \multirow{3}{*}{4090} & BF16 & 1.40 & 1.44 & 1.51 & 1.74 \\
 & & INT4 & 1.70 (0.82$\times$) & 1.81 (0.80$\times$) & 1.81 (0.83$\times$) & 1.84 (0.94$\times$) \\
 & & INT4-Rot & 2.00 (0.70$\times$) & 2.12 (0.68$\times$) & 2.11 (0.72$\times$) & 2.12 (0.82$\times$) \\
 \cmidrule(lr){2-7}
 & \multirow{2}{*}{5090} & BF16 & 0.70 & 0.82 & 1.77 & 2.81 \\
 & & NVFP4 & 1.87 (0.37$\times$) & 1.89 (0.43$\times$) & 2.50 (0.71$\times$) & 3.51 (0.80$\times$) \\
\bottomrule
\end{tabular}
\caption{Inference latency and speedup comparison under different batch sizes on LLaMA-3.1-8B. Rot represents applying online rotation to the down-projection's input in the MLP. }
\label{tab:inference_speed}
\end{table}

As shown in Table~\ref{tab:inference_speed}, W4A4 quantization excels primarily in the prefill phase and under larger batch sizes, achieving approximately a 2$\times$ speedup over BF16 during prefill. In the decode phase with small batch sizes, W4A4 shows no distinct advantage over BF16 due to memory-bound constraints. However, as the batch size increases and the workload becomes compute-bound, the latency gap narrows and eventually reverses, allowing W4A4 to fully exploit its low-precision computational speedup.

\subsubsection{Method inference overhead analysis}
Table \ref{tab:cost-analysis} presents the theoretical analysis of inference cost for the evaluated methods. For experiments in Table \ref{tab:experiment1} and \ref{tab:experiment2}, we only apply online rotation to the down-projection's input in the MLP in each block. We measured the additional latency overhead introduced by this online Hadamard rotation compared to INT4 without rotation. As shown in Table \ref{tab:inference_speed}, this overhead is minimal, while the corresponding performance improvement is substantial, yielding a highly favorable cost-effectiveness ratio.


Besides, the overhead of low-rank compensation is the most noteworthy aspect to discuss. As the hardware implementation largely affects this implementation latency and LQER~\cite{zhang2024lqer} shows that low-rank compensation is hardware-friendly through circuit area analysis, we only compute the additional memory cost and computation (FLOPs) of the low-rank computation, as shown in Table \ref{tab:cost-analysis}. 

In our experiments, we set $r=32$. For LLaMA-3.2-1B, the low-rank compensation introduces approximately 1.41 M additional parameters per decoder layer, compared to about 61 M parameters in all linear layers, costing roughly 2.3\% additional memory overhead in FP16. After quantization, it costs roughly 9.2\% additional memory overhead. And it has a computation overhead ratio of $3.1\%$ in FP16.

\begin{table}[ht]
\centering
\renewcommand{\arraystretch}{0.8}
\begin{tabular}{lll}
\toprule
\textbf{Method} & \textbf{Implementation} & \textbf{Inference Overhead} \\
\midrule
Scaling & Fused into weight & None \\
\midrule
\multirow{2}*{Rotation} & Most: Fused into weight&Fused rotation: None \\
& Few: Online Hadamard rotation & Online rotation: Fast Hadamard\\
\midrule
GPTQ & 
Directly weight changed & 
None \\
\midrule
\multirow{4}*{Low-rank}&  Additional memory: $ \begin{aligned}[t]& A \in \mathbb{R}^{c_{\text{in}} \times r}\\& B \in \mathbb{R}^{r \times c_{\text{out}}}\end{aligned}$ & Memory: $\begin{aligned}[t]\text{ratio} =\frac{r \times (c_{\text{in}} + c_{\text{out}})}{c_{\text{in}} c_{\text{out}}} \end{aligned}$ \\ 
& Additional computation: $XAB$  & Computation: 
$ \begin{aligned}[t] \text{ratio} = \frac{r \times (c_{\text{in}} + c_{\text{out}})}{c_{\text{in}} \times c_{\text{out}}}
\end{aligned} $ \\
\bottomrule
\end{tabular}
\caption{Inference Cost Analysis.}\label{tab:cost-analysis}
\end{table}


\begin{table}[h]
\centering
\renewcommand{\arraystretch}{0.8}
\begin{tabular}{lc}
\toprule
Component & Additional Cost  \\
\midrule
Opt. Rot+Sca & Training: 7 \textasciitilde 8 min; Peak Training GPU: 27.53 GB  \\
GPTQ & Calibration: 2\textasciitilde 3 min  \\
Low-rank ($k$=32) & Calibration: 1\textasciitilde 2min \\
\bottomrule
\end{tabular}
\caption{Training/Calibration cost of quantization components on LLaMA-3.2-1B.}
\label{tab:cost_effectiveness}
\end{table}
To complement the theoretical overhead analysis, Table~\ref{tab:cost_effectiveness} compares the additional calibration/training cost of each quantization component. Optimized rotation and scaling require moderate training cost but yield substantial gains. GPTQ provides the largest PPL improvement at negligible cost. Low-rank compensation offers only marginal improvement if GPTQ is supplied and brings much more memory overhead; however, when used alone, it takes little time to calibrate.

\section{Discussion and Limitations}


Our evaluation mainly focuses on low-bit PTQ for text-only  LLMs under W4A4 settings. Within this scope, our experiments provide consistent evidence for several observations, including the effectiveness of combining pre-quantization transformations with error mitigation methods, as well as the impact of quantization settings such as granularity and symmetry. 

However, some conclusions may depend on specific quantization configurations. The effectiveness of individual quantization techniques may vary across different numerical formats. Our experiments show that rotation improves INT4 quantization but provides limited benefits for MXFP4 and NVFP4, suggesting that format-aware quantization strategies require further investigation. Thus, the conclusions drawn in this study should be interpreted within the evaluated low-bit quantization settings. 

Several open directions remain for future exploration. Developing more effective quantization strategies for emerging numerical formats, exploring quantization under diverse precision configurations beyond the evaluated W4A4 settings, and validating these observations across broader model architectures, including vision-language and multimodal models, remain promising directions for future research.

\section{Conclusions}

In this paper, we have presented an experimental evaluation of Large language model quantization. We systematically decouple the commonly used quantization into two key steps: pre-quantization transformation and quantization error mitigation. After a comprehensive review of existing approaches, we choose scaling and rotation in the pre-quantization transformation step, GPTQ, and low-rank compensation in the quantization error mitigation step to further conduct experimental evaluations of W4A4 quantization on LlaMA/Qwen models on the same ground. We found that using the optimized rotation and scaling, together with two compensation methods, demonstrates the best performance. 
For quantization settings, asymmetric quantization and smaller group sizes yield better results. Using asymmetric quantization on activations shows more benefits in performance compared with weights. 
For FP4 quantization, we can adopt methods used in INT4 quantization. However, rotation-based pre-quantization transformation yields no observable performance gains for MXFP4 and NVFP4 compared with scaling-based methods. And we find that the format of scaling factors demonstrates a large effect. For the FP8 format scaling factor, FP8 (E4M3) with an additional per-tensor FP32 scaling factor is better than FP8 (E8M0). For small group sizes, using a 4-bit scaling factor for weights is also feasible, albeit with some loss in accuracy. To achieve optimal performance under FP4 quantization, further investigation is required to explore FP4's full potential and limitations.

\appendix
\section*{CRediT authorship contribution statement}
\textbf{Yutong Liu:} Methodology, Software, Writing -  original draft. \textbf{Cairong Zhao:} Supervision, Resources, Project administration, Writing - review editing. \textbf{Guosheng Hu:} Methodology, Supervision,
Writing - review editing.
\section*{Acknowledgements}
This work was supported by National Natural Science Fund of China \\(No. U25A20527, 62473286). This work was also supported by Shanghai Municipal Science and Technology Major Project (No. 2025SHZDZX025G10).

The authors declare that no generative AI or AI-assisted technologies were used in the preparation of this manuscript.
 \bibliographystyle{elsarticle-num} 
 \bibliography{references}

\section{Open-source codebases}

\begin{table}[htbp]
\centering
\renewcommand{\arraystretch}{0.8}
\begin{tabular}{ll}
\toprule
\textbf{Sub-method}  & \textbf{Adapted Base Codebase} \\
\midrule
Scaling & \url{https://github.com/mit-han-lab/smoothquant} \\
Random Rot & \url{https://github.com/spcl/QuaRot} \\
Opt. Rot  & \url{https://github.com/Huawei-Noah/OSTQuant} \\
Opt. Rot+Sca & \url{https://github.com/Huawei-Noah/OSTQuant} \\
GPTQ  & \url{https://github.com/IST-DASLab/gptq} \\
Low-rank  & \url{https://github.com/chengzhang-lqer/LQER} \\
\bottomrule
\end{tabular}
\caption{Open-source codebases adapted and integrated in our evaluation framework.}\label{tab:code_sources}
\end{table}

Table~\ref{tab:code_sources} summarizes the core open-source repositories upon which our framework is built. For components that integrate existing techniques, our implementation is heavily adapted from and extended based on these official codebases to ensure a fair and consistent benchmarking environment. Our complete unified evaluation codebase will be released upon the acceptance of this manuscript.
\end{document}